# A Differentiable 3D Scene Graph Metric via Contextual Hellinger Triplet Geometry

**U-Chae Jun**[1] **Yoonhyuk Choi**[2]* **Jiwoo Kang**[2]*
[1]Department of IT Engineering
[2]Division of Artificial Intelligence Engineering
Sookmyung Women's University, Seoul, South Korea
{wjsdbco,chldbsgur123,jwkang}@sookmyung.ac.kr

## Abstract

3D scene graph prediction is commonly supervised with local object and predicate classification losses. Although effective for slot-wise label prediction, this formulation does not provide a graph-level semantic distance between a predicted scene graph and its ground truth: semantically mild label errors and structurally disruptive relational errors can be treated similarly, and subject-predicate-object facts are not compared as scene-level units. We propose *Contextual Hellinger Triplet Geometry*, a graph-level supervision objective that measures the semantic and structural discrepancy between 3D scene graphs. Our key idea is to reinterpret the conventional classifier outputs of a 3D scene graph predictor as an aligned probabilistic scene graph field, where object and predicate predictions jointly define contextual subject-predicate-object facts. This formulation yields a differentiable metric-based training objective with an efficient factorized implementation and can be applied to existing 3D scene graph prediction architectures without modifying their network design. Experiments on 3DSSG demonstrate that our supervision improves downstream graph-conditioned 3D scene generation, while ablation studies confirm the complementary roles of the proposed graph-level terms and user studies show better graph descriptiveness, relation plausibility, and structural preservation.

## 1 Introduction

3D semantic scene graphs [40] represent a scene as a structured object-relation system, where object instances are encoded as nodes and semantic or spatial relationships are encoded as edges. This representation provides an object-centric abstraction of 3D environments and supports relational reasoning beyond dense point clouds, meshes, or voxelized reconstructions. Since the introduction of 3D semantic scene graph prediction from indoor reconstructions [40], subsequent work has explored incremental RGB-D scene graph construction [42], knowledge-guided relation prediction [45], visual-linguistic semantic assistance [41], multimodal spatial knowledge accumulation [12], heterogeneous graph learning [28], transformer-based graph modeling [27], open-vocabulary 3D scene graph generation [8, 23], object-centric representation learning [17], and edge-centric relational reasoning [29]. These advances have significantly improved the modeling capacity of 3D scene graph systems. In this paper, we study a fundamental question: How to define similarity or distance between two 3D scene graphs?

The standard supervised formulation of 3D scene graph prediction [12, 17, 27–29, 40, 41] is expressed over object instances and directed object-pair slots. A model predicts object labels for nodes and predicate labels for directed object pairs, and the training objective typically supervises these outputs through local object and predicate classification losses. Such losses are effective for slot-wise label prediction and provide a natural measure of local target agreement. At the graph level, however, local target agreement and semantic similarity are related but distinct: two graphs can have identical object and predicate classification error counts relative to a reference while differing in graph-level semantic similarity. Fig. 1 illustrates this gap: graphs (a) and (b) have the same object and predicate classification error counts relative to the reference, yet differ in graph-level semantic similarity. Graph-to-Scene reconstructions [44] further visualize the difference at the 3D scene level.

*Corresponding authors.

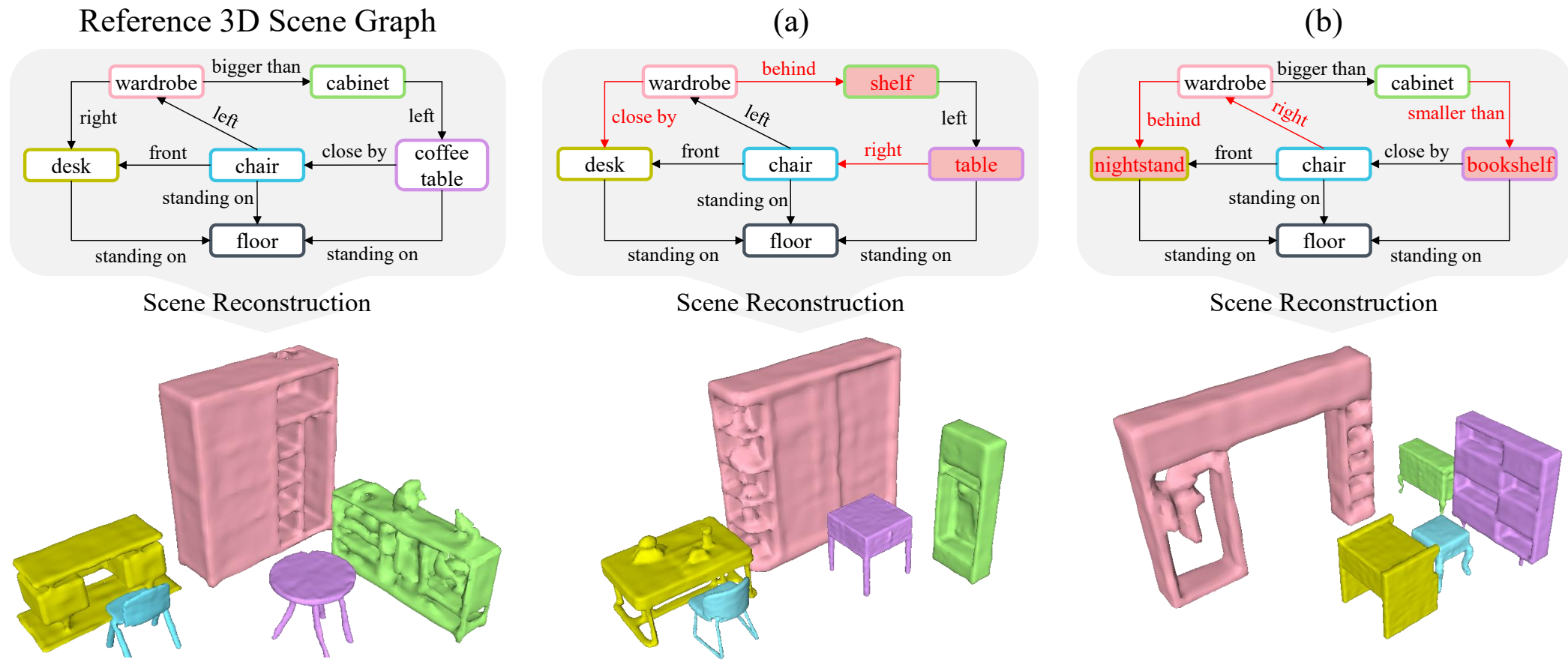

Figure 1: Reference 3D scene graph and two example graphs with identical object and predicate classification error counts but different graph-level semantic similarity to the reference. EchoScene [44] reconstructions visualize this difference at the 3D scene level.

A natural alternative would be to use general graph distance formulations, but they are difficult to use in neural training [31]. Graph edit distance compares graphs through edit operations and is closely related to inexact graph matching [14, 34]. Approximate graph edit distance methods often rely on combinatorial assignment procedures, such as bipartite graph matching [32, 33]. These formulations are suitable for comparing arbitrary graphs [14, 33], but they are not well aligned with supervised 3D scene graph prediction [12, 40, 41]. Without fixed correspondences, graph similarity must first decide which nodes and edges should be matched [14, 33], and the resulting comparison can be ill-posed for scene graph learning because semantically related scenes may differ in topology, indexing, or object decomposition. Moreover, generic edit operations do not directly encode scene-specific semantics, such as object roles, predicate contexts, or subject-predicate-object facts [29, 41, 45].

Our method is inspired by the observation that supervised 3D scene graph prediction provides structure that general graph similarity does not have. In the common object-instance-based setting, the predicted graph and the annotated graph are defined over the same object-instance slots and the same directed object-pair slots. It removes the ambiguity of arbitrary node matching and makes it possible to define a well-specified distance between probabilistic 3D scene graphs.

To this end, we introduce *Contextual Hellinger Triplet Geometry*, a differentiable metric for fixed-correspondence 3D scene graph prediction. The method constructs contextual semantic probability measures for object and relation labels from the training scene graph database. Predicted object and relation distributions are lifted into these contextual measure spaces and compared using Hellinger geometry of probability measures [1, 15]. For each directed object-pair slot, the lifted subject measure, predicate measure, and object measure are composed into a subject-predicate-object product measure. The proposed objective then aligns both corresponding triplet semantics and scene-internal triplet geometry, thereby turning local classifier outputs into a graph-level semantic field.

Our contributions are fourfold. First, we formulate supervised 3D scene graph learning as a differentiable fixed-correspondence graph metric problem, avoiding the correspondence ambiguity of general graph similarity while preserving the relational nature of scene graphs. Second, we introduce contextual semantic probability measures for object and relation labels, including a canonical treatment of no-relation. Third, we define a Hellinger triplet geometry distance that compares subject-predicate-object semantic facts and aligns their scene-internal geometry. Fourth, we show that, under fixed correspondence and identity-anchor assumptions, the square root of the proposed squared loss defines a metric on aligned probabilistic 3D scene graphs, and we provide an exact factorized implementation that avoids explicitly constructing high-dimensional tensor product features.

## 2 Preliminaries and Problem Formulation

**Fixed-Slot 3D Scene Graph Prediction.** We consider supervised 3D scene graph prediction from a 3D scene input $X$, represented by $n$ object instances or segments as graph nodes. We denote the node slot set by

$$V = \{1, \ldots, n\}. \tag{1}$$

For the main formulation, we define relation slots over all directed object pairs,

$$\mathcal{A} = \{(i, j) \mid i, j \in V,\ i \neq j\}, \qquad m = |\mathcal{A}|. \tag{2}$$

Each edge slot $a \in \mathcal{A}$ has a source node $s(a)$ and a target node $t(a)$, so that $a = (s(a), t(a))$. Practical systems may instead use a fixed candidate edge set, but the all-pair formulation provides a clean canonical setting for defining the metric.

Let $\mathcal{C}$ denote the object class vocabulary and let $\mathcal{R}$ denote the positive relation vocabulary. We include an explicit no-relation symbol $\varnothing$ and define

$$\mathcal{R}_0 = \mathcal{R} \cup \{\varnothing\}. \tag{3}$$

A 3D scene graph prediction model produces an object distribution for each node slot,

$$p_i \in \Delta^{|\mathcal{C}|}, \qquad i = 1, \ldots, n, \tag{4}$$

where $p_i(c)$ is the predicted probability that node $i$ belongs to object class $c$. For each directed object-pair slot $a \in \mathcal{A}$, the model produces a relation output. In a multi-class formulation, this output can be written as

$$q_a \in \Delta^{|\mathcal{R}_0|}, \qquad a \in \mathcal{A}. \tag{5}$$

In a multi-label sigmoid formulation, the relation head may instead output positive relation scores

$$s_a \in [0, 1]^{|\mathcal{R}|}. \tag{6}$$

When a sigmoid multi-label head is used, $q_a$ denotes the distribution over $\mathcal{R}_0$ obtained from $s_a$ by the conversion described later in Appendix G.3.

The ground-truth scene graph is defined over the same node and edge slots. Each node slot has an object label

$$y_i^{gt} \in \mathcal{C}, \qquad i = 1, \ldots, n, \tag{7}$$

and each directed object-pair slot has a set of positive relation labels

$$Y_a^{gt} \subseteq \mathcal{R}, \qquad a \in \mathcal{A}. \tag{8}$$

If $Y_a^{gt} = \emptyset$, $a$ is treated as no-relation; prediction and annotation therefore share the same object-instance and object-pair slots, enabling fixed-correspondence metric supervision.

**Local Classification Supervision and Target Objective.** A common abstraction of supervised 3D scene graph learning is an object classification term combined with a predicate classification term. This abstraction underlies the original 3D semantic scene graph formulation [40] and remains a common training backbone in later methods that add visual-linguistic, multimodal, heterogeneous graph, transformer-based, object-centric, or edge-centric reasoning modules [12, 17, 27–29, 41]. The training loss is often written in the form

$$L_{\text{cls}} = \lambda_{\text{obj}} L_{\text{obj}} + L_{\text{pred}}, \tag{9}$$

where $L_{\text{obj}}$ supervises object predictions at node slots and $L_{\text{pred}}$ supervises predicate predictions at edge slots. In the notation above, the object loss can be abstracted as

$$L_{\text{obj}} = \frac{1}{n} \sum_{i=1}^{n} \ell_{\text{obj}} \left(p_i, y_i^{gt}\right), \tag{10}$$

where $\ell_{\text{obj}}$ is typically a cross-entropy loss between the predicted node distribution $p_i$ and the ground-truth node label $y_i^{gt}$. Similarly, the predicate loss can be abstracted as

$$L_{\text{pred}} = \frac{1}{m} \sum_{a \in \mathcal{A}} \ell_{\text{pred}} \left(h_a, Y_a^{gt}\right), \tag{11}$$

where $h_a$ denotes the relation-head output for edge slot $a$. Depending on the formulation, $\ell_{\text{pred}}$ may be a multi-class cross-entropy loss, a binary cross-entropy loss for multi-label predicates, a focal variant, or a related classification loss. The exact form may differ across methods, but the supervision is applied primarily to local object and object-pair targets.

This local classification objective is well suited for learning node and edge classifiers. However, it does not explicitly define a graph-level semantic distance between the predicted graph and the ground-truth graph. The term $L_{\mathrm{obj}}$ compares each predicted node distribution $p_i$ with its ground-truth object label $y_i^{gt}$, and the term $L_{\mathrm{pred}}$ compares each relation output $h_a$ with its ground-truth relation set $Y_a^{gt}$. As a result, the objective decomposes into local node-wise and edge-wise target agreement. It does not directly compare a subject-predicate-object fact as a single semantic unit, nor does it compare the relative geometry among relational facts within the scene. This does not make local classification losses inappropriate; rather, it shows that they optimize a different objective from a scene-level semantic graph distance.

Let the model prediction for scene input $X$ be written as an aligned probabilistic scene graph

$$G_\theta(X) = (\{p_i\}_{i=1}^n, \{q_a\}_{a\in\mathcal{A}}), \tag{12}$$

and let the ground-truth graph be

$$G^{gt} = \left(\{y_i^{gt}\}_{i=1}^n, \{Y_a^{gt}\}_{a\in\mathcal{A}}\right). \tag{13}$$

The conventional local objective and our graph-level objective can be written under the same supervised risk minimization template:

$$\min_\theta \mathbb{E}_{(X,G^{gt})}\left[\mathcal{L}\left(G_\theta(X), G^{gt}\right)\right], \qquad \mathcal{L} = \begin{cases} L_{\mathrm{cls}}, & \text{local classification supervision,} \\ L_{\mathrm{graph}}, & \text{graph-level metric supervision.} \end{cases} \tag{14}$$

Our goal is to replace the purely local classification loss $L_{\mathrm{cls}}$ in Eq. (14) with a differentiable graph-level squared distance $L_{\mathrm{graph}}$ between aligned probabilistic scene graphs. Because $G_\theta(X)$ and $G^{gt}$ are defined over the same object-instance and object-pair slots, this objective does not require arbitrary graph matching. The remaining problem is to design a differentiable fixed-correspondence scene graph distance that reflects contextual semantic similarity among objects, relations, and subject-predicate-object facts. Section 3 introduces such a distance through Contextual Hellinger Triplet Geometry.

# 3 Contextual Hellinger Triplet Geometry

This section defines $L_{\mathrm{graph}}$ for the aligned predicted graph $G_\theta(X)$ and target graph $G^{gt}$ introduced in Section 2, comparing them as probabilistic scene graphs rather than independent local targets.

## 3.1 Contextual Semantic Lifting

We first construct semantic probability measures for object and relation labels from the training scene graph database. For each object class $c \in \mathcal{C}$, we define a contextual object measure

$$\mu_V(c) \in \Delta^{|\Omega_V|}. \tag{15}$$

For each relation label $r \in \mathcal{R}_0$, we define a contextual relation measure

$$\mu_R(r) \in \Delta^{|\Omega_R|}. \tag{16}$$

The context spaces $\Omega_V$ and $\Omega_R$ contain identity and relational-context tokens induced by the training scene graph database [38]. Their full construction is given in Appendix D. The relational part of the object context is built from positive relation occurrences only, so that object semantics are not dominated by absent object-pair relations. For no-relation, we use a canonical absence measure

$$\mu_R(\varnothing) = \delta_{\mathrm{id}:\varnothing}. \tag{17}$$

Using the node prediction in Eq. (4), the predicted object distribution at node slot $i$ is lifted into the object context space as

$$\alpha_i^{pred} = \sum_{c\in\mathcal{C}} p_i(c)\mu_V(c). \tag{18}$$

The corresponding ground-truth object measure is

$$\alpha_i^{gt} = \mu_V(y_i^{gt}), \tag{19}$$

where $y_i^{gt}$ is defined in Eq. (7).

Using the relation prediction in Eq. (5), the predicted relation distribution at edge slot $a$ is lifted as

$$\beta_a^{pred} = \sum_{r \in \mathcal{R}_0} q_a(r)\mu_R(r). \tag{20}$$

The ground-truth relation measure is defined from the relation set in Eq. (8) as

$$\beta_a^{gt} = \begin{cases} \mu_R(\varnothing), & Y_a^{gt} = \emptyset, \\ \frac{1}{|Y_a^{gt}|} \sum_{r \in Y_a^{gt}} \mu_R(r), & Y_a^{gt} \neq \emptyset. \end{cases} \tag{21}$$

If the model uses sigmoid multi-label scores as in Eq. (6), we first convert them into a distribution over $\mathcal{R}_0$ before applying Eq. (20); details are provided in Appendix G.3.

We compare lifted probability measures using Hellinger geometry [1, 15]. For probability measures $P, Q \in \Delta(\Omega)$ over the same finite token space $\Omega$, we define the Bhattacharyya coefficient as

$$\mathrm{BC}(P, Q) = \left\langle \sqrt{P}, \sqrt{Q} \right\rangle = \sum_{\omega \in \Omega} \sqrt{P(\omega)Q(\omega)}. \tag{22}$$

We use the normalized squared Hellinger distance

$$H^2(P, Q) = 1 - \mathrm{BC}(P, Q). \tag{23}$$

This convention follows the standard Bhattacharyya affinity between probability measures [3].

The predicted and ground-truth Hellinger features are defined by

$$v_i^\star = \sqrt{\alpha_i^\star}, \qquad u_a^\star = \sqrt{\beta_a^\star}, \qquad \star \in \{\mathrm{pred}, \mathrm{gt}\}. \tag{24}$$

In Eq. (24), the square root is applied after forming the semantic mixtures in Eqs. (18) and (20). Thus, the comparison is an exact Hellinger comparison between lifted probability measures, rather than a distance between pre-embedded class prototypes.

### 3.2 Subject-Predicate-Object Triplet Representation

Instead of supervising subject, predicate, and object independently, we represent each directed object-pair slot as a contextual subject-predicate-object semantic fact. For an edge slot $a \in \mathcal{A}$ from Eq. (2), the triplet product measure is

$$\rho_a = \alpha_{s(a)} \otimes \beta_a \otimes \alpha_{t(a)}. \tag{25}$$

Its Hellinger feature is

$$z_a = \sqrt{\rho_a} = v_{s(a)} \otimes u_a \otimes v_{t(a)}. \tag{26}$$

Here $\alpha, \beta, v, u$ denote either predicted or ground-truth quantities, depending on the graph under consideration.

For two edge slots $a, b \in \mathcal{A}$, the triplet affinity factorizes as

$$\langle z_a, z_b \rangle = \langle v_{s(a)}, v_{s(b)} \rangle \langle u_a, u_b \rangle \langle v_{t(a)}, v_{t(b)} \rangle. \tag{27}$$

This factorization follows from the product structure in Eq. (25) [4]. It lets us compare subject-predicate-object semantic facts without explicitly constructing the high-dimensional tensor product in Eq. (26).

### 3.3 Graph-Level Distance and Training Objective

Let $G$ and $H$ be two aligned probabilistic scene graphs defined over the same slots as in Eqs. (1) and (2), with $n \geq 2$ so that $m = |\mathcal{A}| > 0$. The aligned triplet term compares corresponding directed object-pair slots:

$$\begin{aligned} D^2_{\mathrm{align}}(G, H) &= \frac{1}{m} \sum_{a \in \mathcal{A}} H^2\left(\rho_a^G, \rho_a^H\right) \\ &= \frac{1}{m} \sum_{a \in \mathcal{A}} \left(1 - \langle z_a^G, z_a^H \rangle\right). \end{aligned} \tag{28}$$

The second equality follows from the squared Hellinger convention in Eq. (23) and the triplet Hellinger feature in Eq. (26). This term measures whether each predicted subject-predicate-object fact is close to its ground-truth counterpart in the contextual Hellinger triplet space.

To compare scene-level relational organization, we first define the triplet Gram matrix

$$K_G[a,b] = \langle z_a^G, z_b^G \rangle, \qquad a,b \in \mathcal{A}. \tag{29}$$

The Gram matrix in Eq. (29) collects pairwise inner products in the triplet Hellinger feature space, following the standard use of Gram matrices in kernel methods [36]. However, our use of this Gram matrix is not merely a generic regularization device. It is equivalent to matching the distance geometry of contextual triplets [16].

To see this, define the triplet distance-geometry matrix of graph $G$ as

$$\Gamma_G[a,b] = H^2\left(\rho_a^G, \rho_b^G\right), \qquad a,b \in \mathcal{A}. \tag{30}$$

By the squared Hellinger convention in Eq. (23) and the triplet Hellinger feature in Eq. (26), the squared Hellinger distance [3, 15] between two triplet product measures is

$$H^2\left(\rho_a^G, \rho_b^G\right) = 1 - \langle z_a^G, z_b^G \rangle. \tag{31}$$

Combining Eqs. (29), (30), and (31), we obtain

$$\Gamma_G[a,b] = 1 - K_G[a,b]. \tag{32}$$

Therefore, for two aligned graphs $G$ and $H$,

$$\left(\Gamma_G[a,b] - \Gamma_H[a,b]\right)^2 = \left(K_G[a,b] - K_H[a,b]\right)^2. \tag{33}$$

Thus, matching triplet Gram matrices is exactly equivalent to matching the scene-internal contextual triplet distance geometry.

The scene-internal triplet geometry term is defined as

$$\begin{aligned} D^2_{\text{geom}}(G,H) &= \frac{1}{m(m-1)} \sum_{\substack{a,b\in\mathcal{A} \\ a\neq b}} \left(\Gamma_G[a,b] - \Gamma_H[a,b]\right)^2 \\ &= \frac{1}{m(m-1)} \sum_{\substack{a,b\in\mathcal{A} \\ a\neq b}} \left(K_G[a,b] - K_H[a,b]\right)^2. \end{aligned} \tag{34}$$

The diagonal is excluded because it corresponds to self-distance or self-affinity and therefore does not encode relative triplet geometry. Unlike the local objective in Eq. (9), Eq. (34) is non-separable at the edge level: each triplet is constrained by its contextual distance to other triplets in the same scene. The geometry term therefore aligns the internal arrangement of semantic relational facts, rather than simply regularizing Gram entries.

The canonical graph distance is

$$D_{\text{metric}}(G,H) = \sqrt{D^2_{\text{align}}(G,H) + D^2_{\text{geom}}(G,H)}. \tag{35}$$

The graph-level supervision term is the squared form of the canonical graph distance:

$$L_{\text{graph}}(G,H) = D^2_{\text{metric}}(G,H) = D^2_{\text{align}}(G,H) + D^2_{\text{geom}}(G,H). \tag{36}$$

In training, Eq. (36) is applied to $G_\theta(X)$ from Eq. (12) and the label-valued target $G^{gt}$ from Eq. (13), after lifting the prediction by Eqs. (18) and (20) and interpreting the target through Eqs. (19) and (21).

For candidate edge settings, some nodes may not appear in any relation slot. Let $E^*$ denote the fixed candidate edge-slot set and let $m^* = |E^*|$. In this variant, the aligned and geometry terms are computed over $E^*$ instead of $\mathcal{A}$, with $m$ replaced by $m^*$. In this case, we use the node-augmented squared distance

$$L^{E^*}_{\text{graph}}(G,H) = D^2_{\text{node}}(G,H) + D^2_{\text{align},E^*}(G,H) + D^2_{\text{geom},E^*}(G,H). \tag{37}$$

Here $D^2_{\text{align},E^*}$ and $D^2_{\text{geom},E^*}$ are defined by Eqs. (28) and (34), respectively, after replacing $\mathcal{A}$ by $E^*$ and $m$ by $m^*$, with $D^2_{\text{align},E^*} = 0$ if $m^* = 0$ and $D^2_{\text{geom},E^*} = 0$ if $m^* < 2$ as in Appendix G.6. The node block is

$$D^2_{\text{node}}(G,H) = \frac{1}{n} \sum_{i=1}^{n} \left(1 - \langle v_i^G, v_i^H \rangle\right). \tag{38}$$

This is not a cross-entropy term; it is the squared form of a Hellinger metric block between lifted object measures and is included only to preserve node information when the active edge set is incomplete.

When the full geometry term in Eq. (34) is expensive and the active off-diagonal pair set is nonempty, we use a uniform sampled estimator. Specifically, $S$ is drawn uniformly without replacement from the active off-diagonal pair set over $\mathcal{E}$, where $\mathcal{E} = \mathcal{A}$ in the all-pair setting and $\mathcal{E} = E^*$ in the candidate edge setting:

$$\widehat{D}^2_{\mathrm{geom}}(G, H) = \frac{1}{|S|} \sum_{(a,b)\in S} \left(K_G[a, b] - K_H[a, b]\right)^2, \qquad \emptyset \neq S \subseteq \{(a, b) \in \mathcal{E}^2 \mid a \neq b\}. \tag{39}$$

This estimator is used only for computational efficiency and does not change the definition of the full all-pair distance in Eq. (35) or the candidate-edge squared objective in Eq. (37). Stratified sampling variants for highly imbalanced no-relation regimes are described in Appendix G.5.

The architecture can remain a standard object and relation classifier. The change is in the supervision: the outputs defined in Eqs. (4) and (5) are structured into a contextual triplet field and optimized through the graph-level objective in Eq. (36), instead of being supervised only through the local classification objective in Eq. (9). In candidate edge settings, Eq. (37) is used instead.

### 3.4 Metric Property and Efficient Computation

We summarize the main theoretical property of Eq. (35) in Theorem 3.1. A full proof is provided in Appendix F.

**Theorem 3.1.** *Assume that the graphs under comparison share the fixed node and edge slots defined in Eqs.* (1) *and* (2)*, that $\mathcal{A}$ contains all directed object pairs with $n \geq 2$, and that the contextual semantic measures in Eqs.* (15) *and* (16) *contain exclusive class-specific identity anchors with positive mass, as specified in Appendix D.3. Also assume the canonical no-relation measure in Eq.* (17)*. Then $D_{\mathrm{metric}}$ in Eq.* (35) *is a metric on aligned probabilistic 3D scene graphs, where graph identity is defined under the fixed node and edge correspondence.*

*Proof (Summary).* The exclusive class-specific identity anchors make the semantic lifting maps in Eqs. (18) and (20) injective; the explicit recovery argument is given in Appendix F.2. The square root of the aligned term in Eq. (28) is an $\ell_2$ product metric of Hellinger distances over corresponding triplet product measures. The square root of the geometry term in Eq. (34) is a Euclidean pseudometric between off-diagonal contextual triplet distance-geometry descriptors, equivalently computed through the Gram representation by Eq. (33). Their direct-sum form in Eq. (35) therefore satisfies non-negativity, symmetry, and the triangle inequality.

If $D_{\mathrm{metric}}(G, H) = 0$, then $D^2_{\mathrm{align}}(G, H) = 0$, so all aligned triplet product measures in Eq. (25) are equal. Taking marginals gives equality of all lifted node and relation measures, and injectivity of the semantic lifting gives equality of the original node and edge distributions. Hence $G = H$ under fixed correspondence.

Finally, Eq. (27) gives an exact implementation of each Gram entry without explicitly constructing $z_a$. Thus, all aligned and geometry terms can be computed from object and relation affinity matrices. The full matrix implementation and complexity analysis are given in Appendix H. □

## 4 Experiments

In this section, we evaluate the effectiveness of the proposed graph-level supervision for 3D scene graph prediction. Sec. 4.1 describes the dataset, metrics, and experimental protocol. Sec. 4.2 evaluates predicted graphs through downstream graph-conditioned 3D scene generation and analyzes the contribution of each graph-level supervision term in Eq. (36). Sec. 4.3 presents user studies on predicted graph quality and generated 3D scene quality. Appendix C.1 provides additional downstream generation results, Appendix C.4 studies the candidate-edge setting, and Appendix C.3 and Appendix C.5 report conventional prediction metrics and failure cases, respectively.

Table 1: Comparison and ablation study of graph-level supervision for graph-conditioned scene generation using EchoScene [44]. All variants use the same OCRL backbone [17] and the same generation protocol. KID scores are reported as $\times 10^{-3}$.

| **Method** | $L_{\text{cls}}$ | $D^2_{\text{align}}$ | $D^2_{\text{geom}}$ | **FID** (↓) | **KID** (↓) |
|---|---|---|---|---|---|
| OCRL [17] | ✓ | – | – | 54.62 ± 0.43 | 3.76 ± 0.08 |
| Ours (w/o $D^2_{\text{geom}}$) | – | ✓ | – | 51.68 ± 0.36 | 3.07 ± 0.05 |
| Ours (w/o $D^2_{\text{align}}$) | – | – | ✓ | 52.84 ± 0.39 | 3.14 ± 0.06 |
| **Ours (Full)** | – | ✓ | ✓ | **49.93** ± 0.32 | **2.78** ± 0.04 |

### 4.1 Experimental Details

**Dataset.** We evaluated on 3DSSG [40], built on 3RScan [39]. Following recent 3D scene graph prediction protocols [17, 41], we used the standard train/validation split with 1,553 indoor scenes, 160 object categories, and 26 predicate types.

**Metrics and Task Description.** Because our objective targets graph-level semantic similarity, the primary downstream evaluation is graph-conditioned scene generation: predicted graphs are fed to the same generator, and generated 3D scenes are compared with references using multi-view FID [18], KID [5], and user studies over descriptiveness, relation plausibility, faithfulness, and structural preservation. Training and evaluation details are in Appendices B.1 and B.5.

### 4.2 Graph-Conditioned 3D Scene Generation Evaluation

To evaluate predicted 3D scene graphs from a downstream generation perspective, we used OCRL [17] as the 3D scene graph prediction backbone and EchoScene [44] as the graph-conditioned 3D scene generator. The OCRL baseline is trained with the original local classification objective $L_{\text{cls}}$ in Eq. (9), whereas our variants are trained with the proposed graph-level supervision terms in Eq. (36). For all methods, predicted graphs are decoded using the same rule and passed to the same generator under an identical generation protocol. Following Appendix B.5, we compute FID [18] and KID [5] between multi-view renderings of 3D scenes generated from ground-truth graphs and those generated from predicted graphs. The quantitative comparison and ablation results are summarized in Table 1. Additional comparisons across different prediction backbones and graph-conditioned generators are provided in Appendix C.1, and the candidate-edge setting is studied in Appendix C.4.

**Graph-Conditioned Reconstruction Results.** Our full model achieves the best performance among the compared variants in both FID and KID. Because the same scene generator, graph decoding rule, and rendering-based evaluation protocol are used for all methods, the improvement indicates that the proposed supervision produces predicted graphs that are more effective as conditions for downstream 3D scene generation. In particular, our method better preserves the semantic composition and spatial organization required to generate plausible 3D indoor scenes, rather than optimizing only independent object and predicate predictions. The qualitative results in Figure 2 confirm these findings: scenes generated from our predicted graphs exhibit more plausible object composition and relational layout, and are closer to the scenes generated from ground-truth graphs than those generated from the OCRL predictions.

**Ablation Study.** We further analyze the contribution of the two graph-level terms in Eq. (36). The variant using only the aligned triplet term $D^2_{\text{align}}$ improves over the local classification baseline, suggesting that matching subject-predicate-object facts in the contextual triplet space provides a more scene-aware semantic training signal than independent local classification. The variant using only the scene-internal triplet geometry term $D^2_{\text{geom}}$ also improves the generation metrics, indicating that preserving the relative geometry among relational facts helps maintain scene-level structural coherence. The full model performs best, demonstrating that the two terms are complementary: $D^2_{\text{align}}$ encourages aligned semantic faithfulness, while $D^2_{\text{geom}}$ regularizes the internal relational organization of the predicted graph.

### 4.3 User Study

To complement the distribution-level quantitative evaluation in Sec. 4.2, we conducted an anonymized and randomized A/B-style user study following prior graph-to-3D and indoor 3D scene genera-tion evaluations [11, 37, 43]. We compared the OCRL baseline [17] with our variant built on the same prediction backbone, using EchoScene [44] as the shared graph-conditioned 3D scene generator. Participants completed a randomized pairwise preference study with two groups of questions: *scene-graph comparison*, which asks participants to judge predicted graph quality, and *scene-scene comparison*, which asks them to judge the corresponding generated 3D scene quality. For each pair produced from the same input 3D scene, participants selected A, B, or tie. We recruited 32 participants, each evaluating 30 pairs per task. Further details are provided in Appendix B.6, and the results are presented in Table 2.

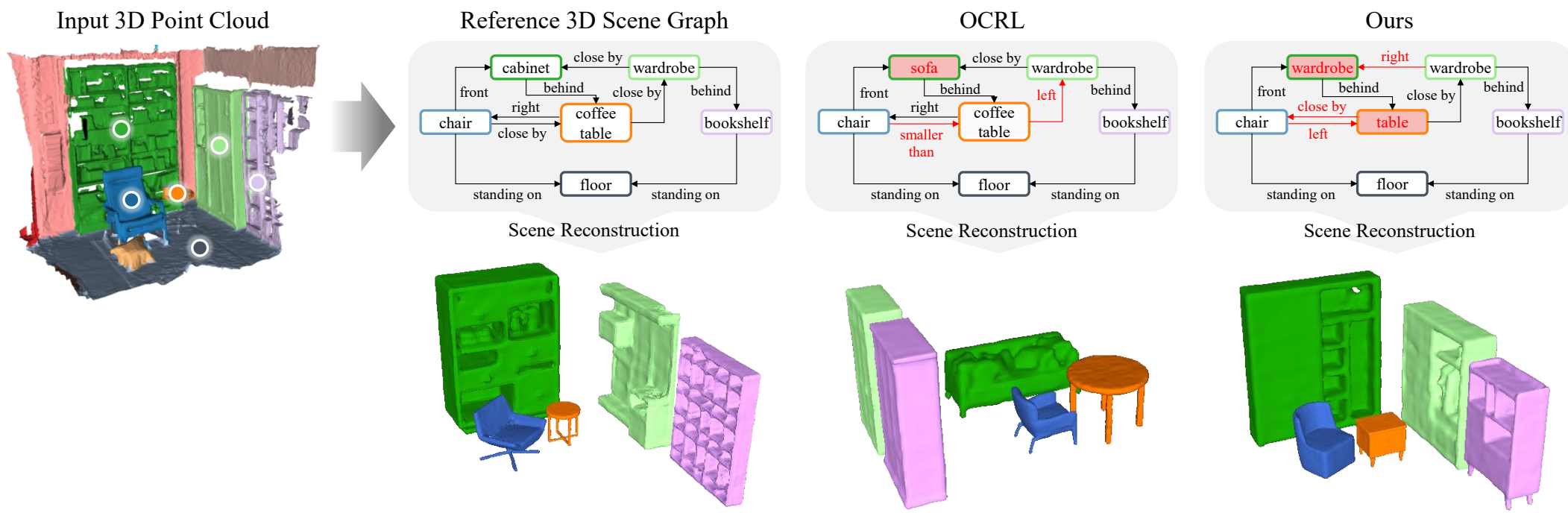

Figure 2: Qualitative results of 3D scene graph predictions and their corresponding 3D scenes generated via EchoScene [44]. The scenes are reconstructed using three different inputs: the ground-truth graph in 3DSSG [40], graphs estimated by OCRL [17], and our method.

Table 2: User study results on scene-graph prediction and graph-conditioned scene generation. Participants compared the baseline (*i.e.*, OCRL [17]) and ours under two questions for each setting.

| Setting | Question | OCRL (%) | Ours (%) | Tie (%) |
|---|---|---|---|---|
| Scene-Graph comparison | Scene-level descriptiveness | 24.1 | 57.7 | 18.2 |
| | Plausibility of object relations | 21.7 | 61.5 | 16.9 |
| Scene-Scene comparison | Faithfulness to the input scene | 26.4 | 54.4 | 19.3 |
| | Structural preservation | 23.3 | 59.4 | 17.3 |

**Scene-Graph Comparison.** The first task evaluates whether the predicted 3D scene graph better captures the semantic content and relational structure of the input 3D scene. For each input scene, participants compared two predicted graphs without knowing the method identity and judged them according to *scene-level descriptiveness* and *plausibility of object relations*. Across both criteria, participants consistently preferred our method over the baseline, indicating that the proposed graph-level supervision produces scene graphs that better describe the overall scene and encode more plausible object relations. This result supports our motivation that training with graph-level semantic and structural consistency can improve the perceptual quality of the predicted graph beyond local object and predicate classification.

**Scene-Scene Comparison.** The second task evaluates whether the generated 3D scene conditioned on the predicted graph better preserves the input scene. For each input scene, participants compared two generated scenes obtained by feeding the predicted graphs into the same graph-conditioned 3D scene generator, and evaluated *faithfulness to the input scene* and *structural preservation*. The consistent preference for our method suggests that our graph predictions provide more effective conditions for downstream 3D scene generation. In particular, the preference in structural preservation suggests that the proposed supervision helps retain scene-level spatial organization, which is difficult to capture using only independent local prediction objectives.

## 5 Conclusion

We presented *Contextual Hellinger Triplet Geometry*, a graph-level supervision objective for 3D scene graph prediction. By reinterpreting conventional object and predicate classifier outputs as an aligned probabilistic scene graph field, our method compares subject-predicate-object facts and their scene-internal relational geometry through an efficient differentiable metric-based objective.

Experiments on 3DSSG show that the proposed supervision better preserves scene-level semantics and spatial organization, improving predicted graphs as conditions for graph-conditioned 3D scene generation. While the current formulation focuses on supervised instance-level prediction with contextual statistics from training annotations, future work may extend it to partially matched, open-vocabulary, or dynamically constructed scene graphs.

## References

[1] Shun-ichi Amari and Hiroshi Nagaoka. *Methods of Information Geometry*, volume 191 of *Translations of Mathematical Monographs*. American Mathematical Society, 2000.

[2] Nathan Aronszajn. Theory of reproducing kernels. *Transactions of the American Mathematical Society*, 68(3):337–404, 1950.

[3] Anil Bhattacharyya. On a measure of divergence between two statistical populations defined by their probability distributions. *Bulletin of the Calcutta Mathematical Society*, 35:99–109, 1943.

[4] Patrick Billingsley. *Probability and Measure*. Wiley, New York, 3 edition, 1995.

[5] Mikołaj Bińkowski, Danica J Sutherland, Michael Arbel, and Arthur Gretton. Demystifying mmd gans. In *Proceedings of the International Conference on Learning Representations*, 2018.

[6] Ingwer Borg and Patrick JF Groenen. *Modern multidimensional scaling: Theory and applications*. Springer, New York, 2005.

[7] Horst Bunke. On a relation between graph edit distance and maximum common subgraph. *Pattern Recognition Letters*, 18(8):689–694, 1997.

[8] Lianggangxu Chen, Xuejiao Wang, Jiale Lu, Shaohui Lin, Changbo Wang, and Gaoqi He. CLIP-driven open-vocabulary 3D scene graph generation via cross-modality contrastive learning. In *Proceedings of the IEEE/CVF Conference on Computer Vision and Pattern Recognition*, pages 27863–27873, 2024.

[9] Tianshui Chen, Weihao Yu, Riquan Chen, and Liang Lin. Knowledge-embedded routing network for scene graph generation. In *Proceedings of the IEEE/CVF Conference on Computer Vision and Pattern Recognition*, pages 6163–6171, 2019.

[10] Donatello Conte, Pasquale Foggia, Carlo Sansone, and Mario Vento. Thirty years of graph matching in pattern recognition. *International Journal of Pattern Recognition and Artificial Intelligence*, 18(3): 265–298, 2004.

[11] Helisa Dhamo, Fabian Manhardt, Nassir Navab, and Federico Tombari. Graph-to-3D: End-to-end generation and manipulation of 3D scenes using scene graphs. In *Proceedings of the IEEE/CVF Conference on Computer Vision and Pattern Recognition*, pages 16352–16361, 2021.

[12] Mingtao Feng, Haoran Hou, Liang Zhang, Zijie Wu, Yulan Guo, and Ajmal Mian. 3D spatial multimodal knowledge accumulation for scene graph prediction in point cloud. In *Proceedings of the IEEE/CVF Conference on Computer Vision and Pattern Recognition*, pages 9182–9191, 2023.

[13] Huan Fu, Bowen Cai, Lin Gao, Ling-Xiao Zhang, Jiaming Wang, Cao Li, Qixun Zeng, Chengyue Sun, Rongfei Jia, Binqiang Zhao, et al. 3D-Front: 3D furnished rooms with layouts and semantics. In *Proceedings of the IEEE/CVF International Conference on Computer Vision*, pages 10933–10942, 2021.

[14] Xinbo Gao, Bing Xiao, Dacheng Tao, and Xuelong Li. A survey of graph edit distance. *Pattern Analysis and Applications*, 13(1):113–129, 2010.

[15] Alison L Gibbs and Francis Edward Su. On choosing and bounding probability metrics. *International Statistical Review*, 70(3):419–435, 2002.

[16] John C Gower. Some distance properties of latent root and vector methods used in multivariate analysis. *Biometrika*, 53(3-4):325–338, 1966.

[17] KunHo Heo, GiHyeon Kim, SuYeon Kim, and MyeongAh Cho. Object-centric representation learning for enhanced 3D semantic scene graph prediction. *Advances in Neural Information Processing Systems*, 2025.

[18] Martin Heusel, Hubert Ramsauer, Thomas Unterthiner, Bernhard Nessler, and Sepp Hochreiter. GANs trained by a two time-scale update rule converge to a local nash equilibrium. *Advances in Neural Information Processing Systems*, 2017.

[19] U Jun, Jaeeun Ko, and Jiwoo Kang. Generative adversarial diffusion. In *Proceedings of the IEEE/CVF International Conference on Computer Vision*, pages 16786–16796, 2025.

[20] U-Chae Jun, Jaeeun Ko, and Jiwoo Kang. Latent diffusion-gan: Adversarial learning in the autoencoded latent space. In *Computer Graphics Forum*, page e70409. Wiley Online Library, 2026.

[21] U-Chae Jun, Jaeeun Ko, and Jiwoo Kang. UniCross3D: Unified cross-view and cross-domain diffusion for consistent single-image 3D generation. In *Computer Graphics Forum*, page e70378. Wiley Online Library, 2026.

[22] U-Chae Jun, Jaeeun Ko, and Jiwoo Kang. View-adaptive renderer for view-consistent 2D-to-3D generation. *Multimedia Systems*, 32(8):511, 2026.

[23] Sebastian Koch, Narunas Vaskevicius, Mirco Colosi, Pedro Hermosilla, and Timo Ropinski. Open3DSG: Open-vocabulary 3D scene graphs from point clouds with queryable objects and open-set relationships. In *Proceedings of the IEEE/CVF Conference on Computer Vision and Pattern Recognition*, pages 14183–14193, 2024.

[24] Nils M Kriege, Fredrik D Johansson, and Christopher Morris. A survey on graph kernels. *Applied Network Science*, 5(1):6, 2020.

[25] Ilya Loshchilov and Frank Hutter. Decoupled weight decay regularization. In *Proceedings of the International Conference on Learning Representations*, 2019.

[26] Cewu Lu, Ranjay Krishna, Michael Bernstein, and Li Fei-Fei. Visual relationship detection with language priors. In *Proceedings of the European Conference on Computer Vision*, pages 852–869. Springer, 2016.

[27] Changsheng Lv, Mengshi Qi, Xia Li, Zhengyuan Yang, and Huadong Ma. SGFormer: Semantic graph transformer for point cloud-based 3D scene graph generation. In *Proceedings of the AAAI Conference on Artificial Intelligence*, pages 4035–4043, 2024.

[28] Yanni Ma, Hao Liu, Yun Pei, and Yulan Guo. Heterogeneous graph learning for scene graph prediction in 3D point clouds. In *Proceedings of the European Conference on Computer Vision*, pages 274–291. Springer, 2024.

[29] Yanni Ma, Hao Liu, Yulan Guo, Theo Gevers, and Martin R Oswald. Edge-centric relational reasoning for 3D scene graph prediction. In *Proceedings of the AAAI Conference on Artificial Intelligence*, pages 7847–7855, 2026.

[30] Anh-Duc Nguyen, Seonghwa Choi, Woojae Kim, Jongyoo Kim, Heeseok Oh, Jiwoo Kang, and Sanghoon Lee. Single-image 3-D reconstruction: Rethinking point cloud deformation. *IEEE Transactions on Neural Networks and Learning Systems*, 35(5):6613–6627, 2022.

[31] Marin Vlastelica Pogančić, Anselm Paulus, Vit Musil, Georg Martius, and Michal Rolinek. Differentiation of blackbox combinatorial solvers. In *Proceedings of the International Conference on Learning Representations*, 2020.

[32] Kaspar Riesen and Horst Bunke. Approximate graph edit distance computation by means of bipartite graph matching. *Image and Vision Computing*, 27(7):950–959, 2009.

[33] Kaspar Riesen, Xiaoyi Jiang, and Horst Bunke. Exact and inexact graph matching: Methodology and applications. In Charu C. Aggarwal and Haixun Wang, editors, *Managing and Mining Graph Data*, volume 40 of *Advances in Database Systems*, pages 217–247. Springer, 2010.

[34] Alberto Sanfeliu and King-Sun Fu. A distance measure between attributed relational graphs for pattern recognition. *IEEE Transactions on Systems, Man, and Cybernetics*, 13(3):353–362, 1983.

[35] Isaac J Schoenberg. Metric spaces and positive definite functions. *Transactions of the American Mathematical Society*, 44(3):522–536, 1938.

[36] Bernhard Schölkopf and Alexander J. Smola. *Learning with Kernels: Support Vector Machines, Regularization, Optimization, and Beyond*. MIT Press, 2002.

[37] Jiapeng Tang, Yinyu Nie, Lev Markhasin, Angela Dai, Justus Thies, and Matthias Nießner. DiffuScene: Denoising diffusion models for generative indoor scene synthesis. In *Proceedings of the IEEE/CVF Conference on Computer Vision and Pattern Recognition*, pages 20507–20518, 2024.

[38] Peter D Turney and Patrick Pantel. From frequency to meaning: Vector space models of semantics. *Journal of Artificial Intelligence Research*, 37:141–188, 2010.

[39] Johanna Wald, Armen Avetisyan, Nassir Navab, Federico Tombari, and Matthias Nießner. RIO: 3D object instance re-localization in changing indoor environments. In *Proceedings of the IEEE/CVF International Conference on Computer Vision*, pages 7658–7667, 2019.

[40] Johanna Wald, Helisa Dhamo, Nassir Navab, and Federico Tombari. Learning 3D semantic scene graphs from 3D indoor reconstructions. In *Proceedings of the IEEE/CVF Conference on Computer Vision and Pattern Recognition*, pages 3961–3970, 2020.

[41] Ziqin Wang, Bowen Cheng, Lichen Zhao, Dong Xu, Yang Tang, and Lu Sheng. VL-SAT: Visual-linguistic semantics assisted training for 3D semantic scene graph prediction in point cloud. In *Proceedings of the IEEE/CVF Conference on Computer Vision and Pattern Recognition*, pages 21560–21569, 2023.

[42] Shun-Cheng Wu, Johanna Wald, Keisuke Tateno, Nassir Navab, and Federico Tombari. SceneGraphFusion: Incremental 3D scene graph prediction from RGB-D sequences. In *Proceedings of the IEEE/CVF Conference on Computer Vision and Pattern Recognition*, pages 7515–7525, 2021.

[43] Guangyao Zhai, Evin Pınar Örnek, Shun-Cheng Wu, Yan Di, Federico Tombari, Nassir Navab, and Benjamin Busam. CommonScenes: Generating commonsense 3D indoor scenes with scene graph diffusion. *Advances in Neural Information Processing Systems*, 36:30026–30038, 2023.

[44] Guangyao Zhai, Evin Pınar Örnek, Dave Zhenyu Chen, Ruotong Liao, Yan Di, Nassir Navab, Federico Tombari, and Benjamin Busam. EchoScene: Indoor scene generation via information echo over scene graph diffusion. In *Proceedings of the European Conference on Computer Vision*, pages 167–184. Springer, 2024.

[45] Shoulong Zhang, Shuai Li, Aimin Hao, and Hong Qin. Knowledge-inspired 3D scene graph prediction in point cloud. *Advances in Neural Information Processing Systems*, 34:18620–18632, 2021.

# Appendix Table of Contents

---

# A Related Work

## A.1 3D Semantic Scene Graph Prediction

3D semantic scene graph prediction has been studied across multiple input modalities and modeling paradigms. Wald et al. [40] established the benchmark formulation of learning 3D semantic scene graphs from indoor reconstructions, and SceneGraphFusion [42] extended the idea to incremental RGB-D observations. For point-cloud-based 3D scene graph prediction, subsequent work improved relational prediction through knowledge-guided reasoning [45], visual-linguistic supervision [41], and multimodal spatial knowledge accumulation [12]. More recent methods further strengthened relational modeling using heterogeneous graph learning [28], transformer-based graph reasoning [27], object-centric representation learning [17], and edge-centric relational reasoning [29].

These methods differ substantially in backbone architecture, feature construction, and auxiliary supervision. However, they largely share the same underlying prediction setting as our work: object predictions are defined on fixed object-instance slots, and relation predictions are defined on directed object-pair slots. Our work is complementary to this literature. Rather than proposing a new 3D scene graph backbone, a new multimodal fusion module, or a new language-guided reasoning component, we focus on the structure of the training objective itself and ask how a predicted graph should be compared to its aligned ground truth at the scene level.

3D scene graphs are also used as structured scene representations beyond recognition alone. For example, graph-conditioned 3D scene generation and diffusion-based synthesis have been studied in CommonScenes [43] and EchoScene [44]. More broadly, diffusion-based generative modeling has also been combined with adversarial learning in both image and autoencoded latent spaces [19, 20]. These downstream uses further motivate supervision that reflects semantic and structural graph similarity, rather than only exact agreement of local object and predicate labels.

### A.2 Graph Similarity and Graph Edit Distance

A natural starting point for graph-level supervision is the literature on graph similarity, graph matching, and graph edit distance. Classical graph edit distance formulations measure dissimilarity through node and edge edit operations on attributed relational graphs [34], and later work clarified its connection to maximum common subgraph formulations [7]. More broadly, graph matching has been studied for several decades as a central problem in structural pattern recognition, where the main challenge is to establish correspondences between graph elements before meaningful comparison can be carried out [10]. This correspondence problem is precisely what makes general graph similarity difficult to use as a differentiable training objective.

For scene graph learning, this difficulty is reflected in the fact that generic graph edit distance is typically combinatorial and expensive to optimize, while approximate solvers often rely on assignment-based relaxations such as bipartite graph matching [32]. Surveys of graph edit distance emphasize both the expressivity and the computational burden of such methods [14]. An alternative family of similarity measures is given by graph kernels, which compare graphs through positive definite similarity functions rather than explicit edit sequences [24]. However, both graph edit distances and graph kernels are primarily designed for comparing arbitrary graphs whose correspondences are not fixed in advance.

Our setting is different. In supervised 3D scene graph prediction, prediction and annotation are defined over the same object-instance slots and directed object-pair slots. Hence, the challenge is not to solve arbitrary graph matching, but to design a differentiable distance on aligned scene graphs that respects scene semantics, contextual subject-predicate-object structure, and scene-internal relational organization. This fixed-correspondence structure places our formulation closer to metric design on aligned probabilistic graph fields than to generic graph matching or arbitrary graph similarity.

### A.3 Hellinger Geometry and Gram Matrices

Our construction is also related to the literature on probability metrics, information geometry, and kernel-based representations. The Hellinger distance is a classical probability metric introduced by Hellinger [1, 15], while the corresponding affinity can be written through the Bhattacharyya coefficient [3]. Modern treatments of probability metrics discuss Hellinger distance alongside other statistical distances and emphasize its favorable geometric and boundedness properties [15]. From the viewpoint of information geometry, the square-root representation maps probability distributions to a spherical geometry, providing a natural Hilbert-space interpretation of Hellinger comparisons [1].

The Gram-matrix side of our method is connected to classical kernel theory. Gram matrices collect pairwise inner products in a feature space, and their role in kernel methods is rooted in reproducing-kernel theory [2] and its practical machine-learning formulation [36]. In our case, the relevant feature space is not a generic kernel embedding but the Hellinger triplet product space induced by contextual subject-predicate-object measures.

Finally, our geometry term is closely related to classical results linking inner-product representations and distance geometry. Schoenberg established a fundamental connection between metric structure and positive definite functions [35], while Gower showed how Euclidean configurations can be recovered from distance information and corresponding Gram-type representations [16]. A modern treatment of this distance-geometry viewpoint, especially through multidimensional scaling and Euclidean configuration recovery from dissimilarity data, is given by Borg and Groenen [6]. These results motivate our use of triplet Gram matrices not merely as regularizers, but as equivalent descriptors of scene-internal contextual triplet distance geometry [6]. In this sense, the proposed geometry term aligns the internal arrangement of semantic relational facts rather than only matching raw pairwise affinities.

## B Additional Implementation Details

### B.1 Detailed Experimental Settings

All experiments follow the standard 3DSSG train/validation split [40]. The prediction backbones are trained using the 3DSSG training set and all reported evaluations are conducted on the held-out 3DSSG validation set. For each compared backbone, we keep the network architecture, input preprocessing, training schedule, and inference procedure fixed, and change only the supervision objective unless otherwise specified. The baseline models are trained with their original local classification objective, while our variants use the proposed graph-level supervision in Eq. (36). For the ablation study, we train variants using $D^2_{\text{align}}$, $D^2_{\text{geom}}$, or both terms.

All models are trained for 100 epochs using the AdamW optimizer [25] with a global batch size of 8 scenes and an initial learning rate of $1 \times 10^{-4}$, decayed with a cosine schedule. The experiments are conducted on two NVIDIA RTX 6000 Ada Generation GPUs with 48GB memory each. The contextual object and relation measures in Eqs. (15) and (16) are computed once from the training split only and then kept fixed throughout training. The main experiments use the all-pair edge setting in Eq. (2); the candidate-edge setting is analyzed separately in Appendix C.4.

For downstream graph-conditioned scene generation, probabilistic predictions are converted into discrete scene graphs using the same decoding rule for all compared methods. Object labels are decoded by the maximum-probability object class,

$$\hat{c}_i = \arg\max_{c \in \mathcal{C}} p_i(c). \tag{40}$$

For each directed object-pair slot $a \in \mathcal{A}$, we first obtain a predicate distribution $q_a \in \Delta^{|\mathcal{R}_0|}$ over $\mathcal{R}_0 = \mathcal{R} \cup \{\varnothing\}$. For sigmoid multi-label relation heads, this distribution is obtained by the conversion in Appendix G.3. We then decode

$$\hat{r}_a = \arg\max_{r \in \mathcal{R}_0} q_a(r). \tag{41}$$

The directed object-pair slot $a$ is emitted as a positive edge only when $\hat{r}_a \neq \varnothing$; otherwise, it is omitted as a no-relation slot. Thus, the no-relation class is used for edge-slot rejection before emitting a positive predicate, rather than being removed before the $\arg\max$. When the downstream generator accepts only one predicate per directed edge, the emitted edge is assigned the single decoded positive predicate $\hat{r}_a$. The decoded graphs are then provided to the selected graph-conditioned scene generator (*i.e.*, CommonScenes [43] or EchoScene [44]) under an identical generation protocol, so that differences in the generated scenes are attributable to the predicted graphs rather than to the generator or evaluation pipeline.

### B.2 Existing Assets, Licenses, and Terms of Use

We use existing datasets, codebases, and pretrained graph-conditioned generators only for research benchmarking and evaluation. We do not redistribute raw 3RScan/3DSSG data or third-party checkpoints; users should obtain these assets from the original sources and comply with their licenses, terms of use, and access restrictions. Table 3 summarizes the existing assets used in this paper.

### B.3 Compute Resources and Reproduction Budget

All reported predictor-training jobs were run on a local workstation equipped with two NVIDIA RTX 6000 Ada Generation GPUs, each with 48GB memory. In the completed runs, all reported configurations fit within the available memory of a single GPU under the two-GPU training setup. We report wall-clock ranges measured from the completed 100-epoch runs used to produce the checkpoints in this paper. Because the exact runtime varies slightly with scene-size composition, validation I/O, and downstream rendering load, we summarize the measurements as ranges rather than as a single number.

Unless otherwise stated, GPU-hours are computed as wall-clock hours multiplied by the number of active GPUs. Predictor training used both GPUs. For downstream generation and rendering, some stages used one GPU and some used both GPUs; therefore, downstream costs are reported directly in accumulated GPU-hours. All GPU-hour numbers should be interpreted as hardware-specific reproduction costs on RTX 6000 Ada hardware, not as hardware-normalized FLOP-hours across GPU generations.

The reproduction budget covers the reported 3D scene graph predictor training, graph-conditioned generation, rendering, Inception feature extraction, and FID/KID evaluation. It assumes that the 3DSSG/3RScan inputs and third-party generator checkpoints have already been obtained and prepared according to the corresponding dataset and codebase instructions. The budget does not include downloading raw datasets, raw 3RScan preprocessing, training baseline-specific external feature extractors from scratch, or training/fine-tuning the external graph-conditioned scene generators. In our evaluation protocol, CommonScenes [43] and EchoScene [44] are used only as fixed pretrained downstream generators.

Table 3: Existing assets used in this paper. For assets whose public repositories or project pages do not specify an explicit reuse license, we use them only for research comparison/evaluation and do not redistribute their code, checkpoints, raw data, or generated benchmark assets.

| **Asset** | **Type** | **Use in this paper** | **License / terms** |
|---|---|---|---|
| 3DSSG [40] | Dataset annotations and 3DSSG documentation / framework reference | Benchmark annotations, train/validation split, object and predicate vocabularies, and data preparation reference. | 3DSSG annotations are distributed from the official 3DSSG project page/GitHub documentation. We do not redistribute the annotations; users should obtain them from the original source. No explicit dataset license was identified in the public 3DSSG documentation we checked. When using the ShunChengWu/3DSSG estimation framework, we follow its CC BY-NC-SA 4.0 license. |
| 3RScan [39] | RGB-D sequences, 3D scans, and reconstructed indoor scenes | Underlying 3D scan data on which 3DSSG is built. | Raw 3RScan/RIO data are obtained through the official access channel after agreeing to the 3RScan Terms of Use. The public 3RScan toolkit repository is MIT-licensed. |
| SGPN / 3DSSG baseline [40] | Baseline method / released framework implementation when applicable | Baseline 3D scene graph prediction comparison. | Credited by citation. When using the released ShunChengWu/3DSSG framework implementation, we follow its CC BY-NC-SA 4.0 license and do not redistribute the original code, checkpoints, or datasets. |
| SGFN [42] | Baseline method / code when applicable | Baseline 3D scene graph prediction comparison. | Credited by citation. The SceneGraphFusion C++ framework is BSD-2-Clause licensed; SGFN-related implementations in the ShunChengWu/3DSSG framework are under that repository's CC BY-NC-SA 4.0 license. We do not redistribute original code or checkpoints. |
| VL-SAT [41] | Baseline method / public research code/checkpoint when applicable | Baseline 3D scene graph prediction comparison. | Credited by citation. The public VL-SAT repository and checkpoint are used only for research comparison when applicable. No explicit reuse license was identified in the public repository; therefore, no third-party code or checkpoint is redistributed in our release. |
| OCRL [17] | Prediction backbone / public research code/checkpoint when applicable | Main prediction backbone and baseline for graph-level supervision experiments. | Credited by citation. The OCRL paper is publicly available under the venue/paper license, but no explicit reuse license was identified for the public code/checkpoint repository. We use the repository only as a research backbone/baseline when applicable and do not redistribute third-party code or checkpoints. |
| CommonScenes [43] | Graph-conditioned 3D scene generator | Downstream generator used to evaluate predicted 3D scene graphs. | Credited by citation. Public research code and provided models are used only as fixed downstream evaluators. No explicit reuse license was identified for the public code/checkpoint repository. The associated SG-FRONT dataset is stated to be MIT-licensed, while 3D-FRONT/3D-FUTURE-derived assets remain subject to their original terms. We do not redistribute third-party code, checkpoints, raw assets, or generated benchmark datasets. |
| EchoScene [44] | Graph-conditioned 3D scene generator | Main downstream generator used to evaluate predicted 3D scene graphs and user-study stimuli. | Credited by citation. Public research code and provided models are used only as fixed downstream evaluators. No explicit reuse license was identified for the public code/checkpoint repository. EchoScene uses SG-FRONT/3D-FUTURE-SDF and optionally 3D-FRONT assets, which remain subject to their original terms. We do not redistribute third-party code, checkpoints, raw assets, or generated benchmark datasets. |

Table 4: Measured predictor-training cost on the two-RTX 6000 Ada workstation. GPU-hours are computed using the two active training GPUs.

| Training group | # runs | Wall-clock / run | GPU-hours / group |
|---|---|---|---|
| Non-OCRL baselines | 3 | 4–6 h | 24–36 |
| OCRL baseline | 1 | 7–12 h | 14–24 |
| Non-OCRL full graph-level variants | 3 | 6–9 h | 36–54 |
| OCRL full graph-level variant | 1 | 9–15 h | 18–30 |
| OCRL all-pair ablations | 2 | 7–15 h | 28–60 |
| OCRL candidate-edge ablations | 2 | 8–13 h | 32–52 |
| Total predictor training | 12 | – | 152–256 |

The contextual object and relation dictionaries in Eqs. (15) and (16) were constructed once from the training split. This preprocessing requires a single pass over the training scenes and took less than 10 minutes on our workstation. Its cost is negligible compared with predictor training and downstream generation.

For the original local classification objective, a single 100-epoch run took 4–6 hours of wall-clock time for the non-OCRL backbones and 7–12 hours for the OCRL-scale backbone on the two-GPU workstation. This corresponds to 8–12 GPU-hours for a non-OCRL run and 14–24 GPU-hours for an OCRL-scale run. With the proposed all-pair graph-level objective, a single 100-epoch run took 6–9 hours for the non-OCRL backbones and 9–15 hours for the OCRL-scale backbone, corresponding to 12–18 and 18–30 GPU-hours, respectively. The increase is consistent with the additional all-pair edge-slot comparisons in the graph-level geometry term analyzed in Appendix H.5, while the shared backbone forward and backward passes still account for a substantial fraction of the runtime.

The OCRL all-pair ablation variants have similar costs in practice. The geometry-only and full graph-level objectives are the most expensive variants because they include edge-slot pair comparisons. The alignment-only variant is asymptotically cheaper than objectives involving $D^2_{\mathrm{geom}}$, but its measured runtime remained within the same broad OCRL-scale range on our hardware. We therefore account for the two OCRL all-pair ablation runs using 7–15 wall-clock hours per run on two RTX 6000 Ada GPUs, or 14–30 GPU-hours per run.

The two candidate-edge ablation variants in Table 8 are also OCRL-scale runs. They are slightly cheaper than the all-pair full objective because the active relation-slot set is restricted from the complete directed edge set $\mathcal{A}$ to the fixed candidate set $E^*$. A 100-epoch candidate-edge run took 8–13 wall-clock hours on the two-GPU workstation, corresponding to 16–26 GPU-hours. The additional node term $D^2_{\mathrm{node}}$ introduced only a small overhead compared with the shared OCRL backbone forward and backward passes, so we use the same range for the candidate-edge variants with and without $D^2_{\mathrm{node}}$.

The reported training configurations consist of four baseline backbone configurations, four corresponding full graph-level configurations, two OCRL all-pair ablation configurations, and two OCRL candidate-edge ablation configurations. The candidate-edge configurations correspond to the variants with and without the node-augmented term $D^2_{\mathrm{node}}$ in Table 8. As summarized in Table 4, reproducing the reported predictor checkpoints requires approximately 155–260 GPU-hours on our RTX 6000 Ada workstation, after rounding the measured range. This includes the predictor checkpoints used in Tables 1, 6, and 8, all trained under the configurations specified in Appendix B.1.

For downstream graph-conditioned scene generation, CommonScenes [43] and EchoScene [44] are used as fixed pretrained generators. The reproduction budget therefore does not include training or fine-tuning these generators from scratch. Following the rendering and FID/KID protocol in Appendix B.5, each generated scene is rendered from five views, and FID/KID are computed from the pooled rendered-image distributions.

For each trained predictor and downstream generator pair, graph conversion, graph-conditioned scene generation, five-view rendering, Inception feature extraction, and FID/KID computation required 1.5–3 additional GPU-hours, depending on the generator and the number of evaluation scenes. There are 20 unique predictor–generator evaluation pairs after counting overlapping OCRL–EchoScene baseline and full-model entries only once: 16 pairs from Table 6, two additional OCRL all-pair ablation pairs from Table 1, and two candidate-edge ablation pairs from Table 8. Thus, the downstream generation, rendering, and FID/KID evaluation stage requires approximately 30–60 GPU-hours. The scene-level bootstrap procedure reuses cached renderings or cached Inception features and adds less than one GPU-hour in total.

Overall, reproducing the reported training and downstream evaluation results requires approximately 185–320 GPU-hours on RTX 6000 Ada hardware. This budget covers the reported predictor training, the two candidate-edge ablation runs, graph-conditioned generation, five-view rendering, and FID/KID evaluation. The broader project involved additional small debugging, pipeline-validation, and sanity-check runs, but the reported results rely only on the configurations listed above.

### B.4 3DSSG-to-Generator Interface for 3D Scene Generation Evaluation

Our generation-based evaluation uses pretrained graph-conditioned 3D scene generators, *i.e.*, CommonScenes [43] and EchoScene [44], as downstream evaluators of predicted 3D scene graphs. These generators are built on a 3D-FRONT-style graph representation [13], whereas 3D scene graph prediction methods are commonly trained and evaluated on 3DSSG [40]. The two graph protocols differ in object vocabulary, predicate vocabulary, object granularity, and supported room-domain coverage. Therefore, raw 3DSSG graphs cannot be directly used as generator inputs without introducing vocabulary and domain mismatch. To avoid confounding graph prediction quality with generator-interface incompatibility, we define a deterministic 3DSSG-to-generator graph conversion protocol.

Importantly, we do not retrain CommonScenes or EchoScene on 3DSSG, nor do we modify their architectures. Instead, both ground-truth and predicted 3DSSG graphs are converted into the generator-compatible graph representation before generation. The conversion rule is fixed before evaluating any prediction model and is applied identically to all compared methods. Thus, the downstream generation results compare the quality of the predicted graphs under a common interface, rather than the ability of a method to exploit a method-specific conversion or filtering rule.

**Deterministic Conversion Operator.** Let $\Phi$ denote the downstream graph-conditioned scene generator. We define a deterministic conversion operator $T_\Phi$ that maps a 3DSSG graph into the graph format accepted by $\Phi$. The operator consists of an object-label map and a predicate-label map. Each 3DSSG object or predicate label is either mapped to a generator-compatible label or assigned to a null symbol $\perp$ when no reliable semantic counterpart exists. Labels mapped to $\perp$ are removed from the converted graph. The mapping is defined at the class-label level and does not depend on the scene, the prediction method, or the prediction confidence.

**Object Label Mapping.** For each object node, we first check whether its 3DSSG object label is directly supported by the generator vocabulary. If so, the label is preserved. For example, generator-compatible furniture categories such as *armchair*, *bookshelf*, *cabinet*, *chair*, and *coffee table* are used without modification. If the exact 3DSSG label is not present but has a clear synonym or super-category in the generator vocabulary, we map it to the corresponding generator label. For instance, *couch* is mapped to *sofa*, *couch table* to *coffee table*, *bedside table* to *nightstand*, and *computer desk* to *desk*. For labels that are represented by a more fine-grained set of generator categories, such as *bed*, we use the canonical category specified by the generator protocol.

Object nodes are removed when their 3DSSG labels have no reliable generator-compatible counterpart. This includes unsupported architectural or structural annotations such as *ceiling*, *door*, *doorframe*, and *pipe*; small clutter objects such as *backpack*, *bag*, *book*, *bottle*, *box*, and *folder*; and facility-specific objects such as *basin*, *bathtub*, *bidet*, *shower*, and *sink*. When an object node is removed, all predicate edges incident to that node are also removed. This step reflects the furniture-centric graph interface of the 3D-FRONT-based generators and prevents unsupported object labels from being passed to the generator.

**Predicate Label Mapping and Edge Cleanup.** After object label mapping, we convert predicate labels between the remaining object nodes. If a 3DSSG predicate label is directly supported by the generator vocabulary, it is preserved. For example, geometric predicates such as *left*, *right*, *front*, and *behind* are retained when supported by the generator. If the 3DSSG predicate label has a clear generator-compatible synonym, we map it to the corresponding predicate label. For example, *higher than* is mapped to *taller than*, *lower than* to *shorter than*, *same symmetry as* to *symmetrical to*, and *supported by* to *standing on*.

Predicates without a reliable layout-level counterpart in the generator vocabulary are removed. Examples include *attached to*, *belonging to*, *build in*, and *connected to*, which are meaningful in 3DSSG annotations but are not part of the generator-compatible predicate interface. No-relation edges are never provided to the generator. If multiple mapped predicate labels remain for the same directed object pair, duplicate labels are merged; when the generator requires a single predicate per directed object pair, we apply the same deterministic edge-collapsing rule to the converted reference graph and to all converted predictions. The conversion does not infer new predicates from 3D coordinates or room layout; it only maps or removes existing graph labels.

**Reference-based Scene Filtering.** We select the generation-evaluation dataset using only the converted reference graphs. For each 3DSSG validation scene $s$, let $G_s^{gt} = (V_s^{gt}, E_s^{gt})$ be the original reference graph and let $\widetilde{G}_s^{gt} = T_\Phi(G_s^{gt}) = (\widetilde{V}_s^{gt}, \widetilde{E}_s^{gt})$ be its converted graph. A scene is included in the generation-evaluation dataset only when the converted reference graph satisfies all of the following conditions:

- the converted graph contains at least five object nodes;
- the converted graph contains at least five positive predicate edges;
- the object node retention ratio is at least $0.5$.

The object node retention ratio is defined as

$$\rho_s = \frac{|\widetilde{V}_s^{gt}|}{|V_s^{gt}|}. \tag{42}$$

This filtering avoids degenerate generator inputs whose object or predicate content is too sparse after conversion. It also prevents scenes whose original 3DSSG content is mostly outside the generator vocabulary from dominating the evaluation. For example, room domains dominated by unsupported facilities or small clutter objects are naturally excluded when their converted reference graphs fail the node-count, predicate-count, or retention-ratio criteria. Conversely, if a generator-compatible indoor scene contains only a small number of unsupported objects, those nodes are removed while the scene remains in the evaluation subset as long as the converted reference graph satisfies the above criteria.

This filtering is performed once using the reference graphs only. Predicted graphs are never used to decide whether a scene is included or excluded. Therefore, all compared methods are evaluated on the same fixed scene dataset, and no method receives an advantage from method-specific scene selection.

**Applying the Protocol to Predicted graphs.** For each selected scene and each prediction method $M$, the decoded predicted graph $\widehat{G}_s^M$ is converted using the same operator $T_\Phi$. Unsupported predicted object labels are removed, unsupported predicted predicate labels are removed, and predicate edges incident to removed object nodes are discarded. The resulting converted prediction $T_\Phi(\widehat{G}_s^M)$ is then provided to the same downstream generator $\Phi$ under the same sampling and rendering protocol. Scenes are not filtered again based on the converted predicted graph.

Under this protocol, the ground-truth graph and all predicted graphs are evaluated through the same generator-compatible interface. The resulting FID/KID scores should therefore be interpreted as generation-based scene-level fidelity. Conventional 3DSSG prediction metrics are reported separately in Appendix C.3. This separation allows us to evaluate two complementary aspects: standard exact-label prediction on the original 3DSSG vocabulary, and downstream 3D scene-level coherence under a controlled graph-to-scene generation interface.

### B.5 Multi-view Rendering-based 3D Scene Generation Evaluation

The standard supervised formulation of 3D scene graph prediction evaluates object, predicate, and subject-predicate-object triplet prediction using recall and mean recall [12, 17, 27–29, 40, 41]. These metrics measure whether each local object or predicate label exactly matches the ground-truth annotation. However, exact local label agreement does not fully capture whether a predicted 3D scene graph preserves the scene-level semantics and relational structure of the ground-truth 3D scene graph. For example, two predicted graphs may have similar recall scores while inducing substantially different global scene layouts when used as conditions for 3D scene generation. Therefore, we report conventional recall-based metrics only as supplementary results in Appendix C.3, and use graph-conditioned 3D scene generation as our main quantitative evaluation.

Our key assumption is that a predicted 3D scene graph is more semantically and structurally faithful to the ground-truth graph if it induces a similar distribution of generated 3D scenes when passed to the same graph-conditioned 3D scene generation model. Before generation, both ground-truth and predicted graphs are converted into the generator-compatible graph representation using the fixed protocol in Appendix B.4. Following recent 3D scene graph-to-3D scene generation methods such as CommonScenes [43] and EchoScene [44], we convert both ground-truth and predicted 3D scene graphs into generated 3D scenes, and compare their rendered distributions using Fréchet Inception Distance (FID) [18] and Kernel Inception Distance (KID) [5]. This evaluation provides a downstream quantitative proxy for how well the predicted graph preserves the semantic composition and spatial organization required for plausible 3D scene generation.

For each validation scene $s$, let $G_s^{gt}$ denote the ground-truth 3D scene graph and $\hat{G}_s^M$ denote the 3D scene graph predicted by method $M$. Using a fixed graph-conditioned 3D scene generator $\Phi$ and the deterministic conversion operator $T_\Phi$, we generate a reference 3D scene $\Phi(T_\Phi(G_s^{gt}))$ and a predicted-graph-conditioned 3D scene $\Phi(T_\Phi(\hat{G}_s^M))$. The same generator, sampling protocol, renderer, and camera configuration are used for all methods. Thus, the evaluation focuses on the quality of the input 3D scene graph rather than differences in the generation or rendering pipeline.

Each generated 3D scene is rendered at $256 \times 256$ resolution from five views: one top-down view and four oblique side views. The four oblique views use azimuth angles of 0°, 90°, 180°, and 270° with a fixed elevation angle of 30°. The camera configuration is determined from the reference scene coordinate frame and reused for the corresponding predicted-graph-conditioned generations, preventing method-specific camera fitting from hiding scale or layout errors. Compared with a single top-down rendering, this multi-view protocol better reflects 3D scene structure, including object arrangement, relative scale, occlusion, and side-view geometric consistency [21, 22].

To reduce nuisance factors that are not directly related to 3D scene graph quality, such as texture, material, and illumination, we use semantic-color rendering. Each object instance is rendered with a category-consistent color, and the same category-to-color mapping is used for all ground-truth-graph-conditioned and predicted-graph-conditioned scenes. This makes the image-based metrics more sensitive to semantic composition and spatial layout rather than photorealistic appearance.

Let $\mathcal{I}_{gt}$ denote the set of multi-view renderings obtained from the 3D scenes generated with ground-truth graphs, and let $\mathcal{I}_M$ denote the set of multi-view renderings obtained from the 3D scenes generated with the predicted graphs of method $M$. For each method, we pool renderings over all validation scenes and all five camera views, and compute FID and KID between $\mathcal{I}_{gt}$ and $\mathcal{I}_M$ using standard Inception features. Lower FID and KID indicate that the predicted graphs induce a generated 3D scene distribution closer to the distribution induced by ground-truth 3D scene graphs. The $\pm$ values reported in the FID/KID tables denote scene-level bootstrap standard errors computed with $B = 1000$ resamples of the generator-compatible validation scenes. Each bootstrap resample samples scenes with replacement and keeps all five rendered views of a sampled scene together, so camera views are not treated as independent validation samples.

Although FID and KID do not directly measure pairwise graph correctness or individual predicate accuracy, they are appropriate for this setting because graph-conditioned 3D scene generation is inherently multimodal: multiple valid 3D scenes can correspond to the same scene graph. This contrasts with single-image 3D reconstruction settings that explicitly target geometric recovery from a visual observation [30]. Therefore, an exact pairwise 3D reconstruction metric would be overly restrictive and would not reflect the distributional nature of the task. We use FID and KID as distribution-level quantitative evidence for downstream 3D scene generation quality, and complement them with the user studies in Sec. 4.3 and Appendix B.6.

### B.6 User Study Design and Protocol

We conducted a user study to evaluate whether the proposed graph-level supervision improves the perceptual quality of predicted 3D scene graphs and their downstream graph-conditioned 3D scene generation results. The study was designed as a pairwise preference test between the baseline method, *i.e.*, OCRL [17], and our method. Our protocol is motivated by prior perceptual evaluations in graph-to-3D and indoor 3D scene generation, which commonly use anonymized or randomized comparative displays and perceptual criteria related to scene realism, condition consistency, graph correctness, and structural plausibility [11, 37, 43]. We use pairwise comparison rather than absolute rating because the quality of a 3D scene graph or a generated 3D scene can be difficult to calibrate across participants with a fixed numerical scale. Pairwise comparison reduces scale bias and directly measures which method is preferred under the same input 3D scene.

**Participants.** We recruited 32 participants, consisting of 16 female and 16 male participants aged between 24 and 32. The participants were students or researchers with backgrounds in computer vision, computer graphics, or related fields. All participants had normal or corrected-to-normal vision and had no prior exposure to the compared methods or the evaluated results.

**Stimuli.** The study was organized into two response blocks: *scene-graph comparison* and *scene-scene comparison*. For both response blocks, we sampled the same 30 scenes from the 3DSSG validation split. For each sampled scene, we constructed one pairwise comparison trial between the OCRL baseline and our method. Each trial displayed the input 3D scene, two predicted 3D scene graphs, and the two corresponding graph-conditioned generated 3D scenes. Thus, each participant evaluated 30 comparison trials, yielding 30 scene-graph comparisons and 30 scene-scene comparisons.

In the scene-graph comparison block, participants judged which predicted graph better described the input 3D scene. The input 3D scene was visualized using the same rendering style for all methods. Each predicted graph was visualized with object category labels as nodes and positive predicate labels as directed edges; no-relation edges were omitted for readability. The same graph decoding rule, node display format, font size, and edge visualization rule were used for both methods.

In the scene-scene comparison block, participants judged which generated 3D scene better preserved the input 3D scene. The generated scenes were obtained by feeding the predicted graphs of baseline and our method into the same graph-conditioned 3D scene generator, EchoScene [44]. Before generation, the predicted graphs were converted using the same deterministic 3DSSG-to-generator protocol described in Appendix B.4. The generator, sampling procedure, renderer, camera configuration, and visualization style were fixed across the two methods. The generated 3D scenes were shown with the same semantic-color rendering style and fixed camera configuration, so that participants could compare object composition, relative positions, and structural consistency without method-specific visualization differences.

**Procedure and Controls.** The study was conducted using a controlled web-based interface. For each comparison trial, the two alternatives were anonymized as Program A and Program B, and participants were not informed which result was produced by which method. The assignment of the baseline and our method to Program A/Program B, corresponding to the left/right columns in the interface, was randomized independently for each participant and each stimulus to reduce positional bias. The order of comparison pairs was also randomized for each participant. The ground-truth scene graph was not shown to participants; participants made their judgments based on the input 3D scene and the two compared outputs. A tie option was provided to avoid forcing a preference when the two results were perceptually similar or equally inadequate.

Participants were instructed to judge the scene-graph criteria from the predicted graphs and the input scene, and to judge the scene-scene criteria from the generated scenes and the input scene. Consistent with the generator-compatible evaluation protocol, the scene-scene judgments focused on object composition and spatial layout rather than texture, illumination, wall geometry, or unsupported clutter.

The exact instruction text shown to participants before the trials was as follows:

- You will compare the outputs of two anonymized programs, Program A and Program B, for the same input 3D scene.

- The identities of the programs are hidden and the left/right order is randomized.
- For the scene-graph questions, please compare the predicted 3D scene graphs with respect to how well they describe the input scene and how plausible their object relations are.
- For the scene-scene questions, please compare the generated 3D scenes with respect to faithfulness to the input scene and preservation of object arrangement, relative positions, and overall 3D structure.
- Please focus on semantic object composition and spatial layout rather than texture, lighting, or rendering style.
- For each question, choose Program A, Program B, or tie if the two results are equally good, equally poor, or not distinguishable.

Figure 3 illustrates an example comparison trial. The interface first presents the common input 3D scene, then displays the anonymized outputs from Program A and Program B in two columns, where each column contains a predicted 3D scene graph and the corresponding graph-conditioned generated 3D scene. The response panels collect A/B/tie preferences for the scene-graph and scene-scene criteria.

**Participant Risk and Ethics.** The study involved only low-risk visual preference judgments over rendered 3D scenes and predicted 3D scene graphs. It did not involve physical intervention, biometric measurement, clinical or psychological questions, deception beyond anonymizing method identities, or collection of sensitive personal information. Participants were informed that their responses would be analyzed only in aggregate and that they could stop participating at any time.

**Questions.** Participants answered two questions for each response block, as summarized in Table 5. The same question wording was used in the response panels of the web interface shown in Figure 3. The questions were designed by adapting perceptual factors used in prior 3D scene generation studies, such as global correctness, realism, graph-constraint correctness, condition consistency, and functional or structural plausibility, to our two evaluation targets: predicted 3D scene graphs and their downstream generated 3D scenes [11, 37, 43]. Specifically, the scene-graph comparison questions evaluate graph-level semantic quality and relation plausibility, while the scene-scene comparison questions evaluate generated scene faithfulness and 3D structural preservation.

Table 5: Exact questions used in the user study. For each question, participants selected A, B, or tie.

| **Task** | **Criterion** | **Question** |
|---|---|---|
| Scene-graph comparison | Scene-level descriptiveness | Which predicted 3D scene graph better describes the input 3D scene as a whole? |
| | Plausibility of object relations | Which predicted 3D scene graph contains more plausible object relations for the input 3D scene? |
| Scene-scene comparison | Faithfulness to the input scene | Which generated 3D scene is more faithful to the input 3D scene? |
| | Structural preservation | Which generated 3D scene better preserves the object arrangement, relative positions, and overall 3D structure of the input 3D scene? |

**Response Aggregation.** For each question, participants selected one of three visible responses: Program A, Program B, or tie. After data collection, Program A/B responses were mapped back to baseline or ours using the hidden randomized assignment for the corresponding trial. Let $N_q^{\mathrm{base}}$, $N_q^{\mathrm{ours}}$, and $N_q^{\mathrm{tie}}$ denote the number of responses assigned to the baseline, our method, and tie for question $q$, respectively. We report the preference rates as

$$P_q^m = 100 \times \frac{N_q^m}{N_q^{\mathrm{base}} + N_q^{\mathrm{ours}} + N_q^{\mathrm{tie}}}, \quad m \in \{\mathrm{base}, \mathrm{ours}, \mathrm{tie}\}. \tag{43}$$

Since 32 participants evaluated 30 pairs for each task, each question received 960 responses. This response count is comparable to prior 3D scene generation user studies, such as Graph-to-3D, which used 20 participants evaluating approximately 30 scenes, and DiffuScene, which collected responses from 45 users over 225 scenes [11, 37]. The final user study results are reported as the preference percentages of baseline, our method, and tie in Table 2.

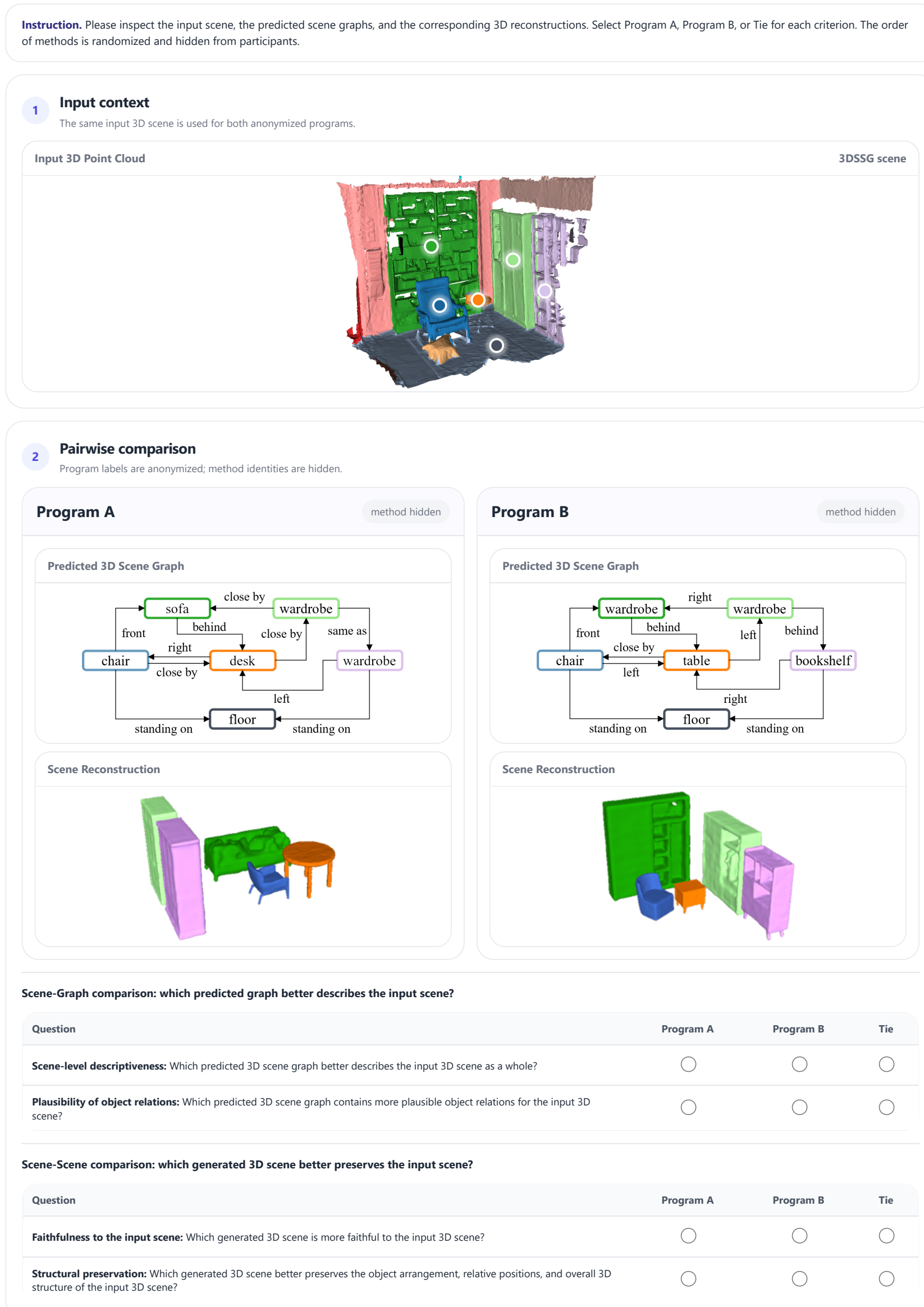

Figure 3: Example web interface used in the pairwise user study. For each input 3DSSG scene, participants compared two anonymized method outputs, Program A and Program B, each containing a predicted 3D scene graph and the corresponding graph-conditioned generated 3D scene. The interface collected A/B/tie preferences for scene-level descriptiveness, object-relation plausibility, input-scene faithfulness, and structural preservation, with method identities hidden and the A/B assignment randomized.

# C Additional Experiments

## C.1 Generalization across Scene Graph Backbones and Generators

We further examine whether the proposed graph-level supervision is tied to a particular 3D scene graph prediction backbone or to a particular graph-conditioned scene generator. To this end, we evaluate four 3D scene graph prediction backbones, SGPN [40], SGFN [42], VL-SAT [41], and OCRL [17], with two downstream graph-conditioned 3D scene generators, CommonScenes [43] and EchoScene [44].

These backbones cover a broad range of 3D scene graph prediction paradigms, from PointNet/GCN-based prediction and feature-fusion-based graph construction to visual-linguistic assisted training and cross-modal contrastive object representation learning. For each backbone, the baseline is trained with its original local object and predicate classification objective, while the corresponding "+ Ours" variant is trained with the proposed graph-level supervision under the same network architecture, graph decoding rule, and downstream generation protocol. Thus, each baseline-to-ours comparison isolates the effect of the supervision objective while keeping the prediction model and generation pipeline fixed. More detailed experimental settings are provided in Appendix B.1. Following the protocol in Appendix B.5, we compare the distribution of multi-view renderings of scenes generated from ground-truth graphs with that of scenes generated from predicted graphs. Thus, lower FID and KID indicate that the predicted graphs induce generated scenes closer to those conditioned on ground-truth scene graphs.

The quantitative results are summarized in Table 6, and qualitative comparisons are shown in Figure 4. Here, CommonScenes and EchoScene are used as fixed downstream evaluators of predicted graph quality. The reported FID/KID values are therefore not intended to be directly compared with the original FID/KID numbers reported by those generators under their own benchmarks.

**Backbone-Agnostic Improvement.** Across all prediction backbones and both downstream generators, applying our graph-level supervision consistently improves both FID and KID. With CommonScenes, the proposed supervision reduces FID by $4.44$–$6.02$ and KID by $1.04$–$1.08$ in the $\times 10^{-3}$ scale. With EchoScene, it reduces FID by $4.69$–$5.92$ and KID by $0.88$–$1.13$ in the same scale. This consistent improvement across SGPN, SGFN, VL-SAT, and OCRL shows that the proposed objective is not specialized to a single backbone architecture.

The trend across backbones is also broadly consistent with the relative strength of the underlying 3D scene graph predictors. Backbones with stronger conventional prediction performance generally yield lower generation errors, and this ordering is preserved after applying our supervision. Importantly, our method improves both earlier backbones such as SGPN and SGFN and stronger recent backbones such as VL-SAT and OCRL. This suggests that the proposed graph-level objective is complementary to backbone-level representation learning. Rather than relying on a particular feature extractor or relation classifier, it provides an additional supervision signal that makes the predicted graph more useful as a structured condition for downstream scene generation.

**Generator-Agnostic Robustness.** The same improvement pattern is observed with both CommonScenes and EchoScene. Although the two generators have different graph-conditioning and generation mechanisms, the relative advantage of our method remains stable. Under this protocol, EchoScene yields lower absolute FID and KID values than CommonScenes for the same predicted-graph inputs; however, the key observation is that the baseline-to-ours improvement is preserved for every backbone in both generator settings. This indicates that the gains are not an artifact of a specific generator architecture, sampling behavior, or conditioning interface.

This robustness is important because graph-conditioned scene generation evaluates the predicted graph through a downstream consumer. If the improvement were observed only with a single generator, it could be attributed to generator-specific biases. In contrast, the agreement between CommonScenes and EchoScene suggests that the gains originate from improved predicted-graph conditions rather than from a generator-specific artifact: the generated scenes become closer to the ground-truth-graph-conditioned scenes regardless of which downstream generator consumes the graph.

**Qualitative Analysis.** The qualitative comparisons complement the quantitative results by showing how graph-level prediction errors affect downstream scene generation. In the reference example, the scene graph specifies a coherent indoor arrangement involving a desk, chair, sofa, table, floor, and storage furniture, together with support, proximity, directional, and comparative relations such as *standing on*, *close by*, *left/right/front*, and *higher than*. Baseline predictors often produce locally plausible but globally inconsistent graph reconstructions, including substituted object categories, duplicated furniture roles, and inconsistent spatial or comparative relations. These graph-level inconsistencies are reflected in the generated scenes as missing or extraneous furniture, unstable object placement, and layouts that deviate from the reference-graph-conditioned reconstructions.

Table 6: Comparison of graph-conditioned scene generation quality across different downstream generators. Lower values indicate better generation quality. KID scores are reported as $\times 10^{-3}$.

| Method | CommonScenes [43] | | EchoScene [44] | |
|---|---|---|---|---|
| | **FID** (↓) | **KID** (↓) | **FID** (↓) | **KID** (↓) |
| SGPN [40] | 70.95 ± 0.65 | 6.92 ± 0.13 | 66.74 ± 0.60 | 5.82 ± 0.11 |
| SGPN [40] + Ours | **66.18** ± 0.54 | **5.88** ± 0.10 | **62.04** ± 0.49 | **4.94** ± 0.09 |
| SGFN [42] | 67.76 ± 0.59 | 6.31 ± 0.12 | 63.48 ± 0.54 | 5.31 ± 0.10 |
| SGFN [42] + Ours | **61.92** ± 0.48 | **5.24** ± 0.09 | **58.11** ± 0.43 | **4.36** ± 0.08 |
| VL-SAT [41] | 62.10 ± 0.50 | 5.10 ± 0.09 | 57.84 ± 0.45 | 4.18 ± 0.08 |
| VL-SAT [41] + Ours | **56.08** ± 0.40 | **4.02** ± 0.07 | **51.92** ± 0.36 | **3.05** ± 0.05 |
| OCRL [17] | 58.64 ± 0.47 | 4.70 ± 0.09 | 54.62 ± 0.43 | 3.76 ± 0.08 |
| OCRL [17] + Ours | **54.20** ± 0.35 | **3.64** ± 0.06 | **49.93** ± 0.32 | **2.78** ± 0.04 |

In contrast, the corresponding "+ Ours" variants more reliably preserve the relational scaffold of the scene. Even when individual object labels are not perfectly recovered, the predicted graphs better maintain support relations to the floor, lateral ordering around the sofa, proximity between functionally related objects, and the relative placement of large storage furniture. Consequently, both CommonScenes and EchoScene generate scenes with clearer object composition and more stable spatial organization. This visual trend appears not only for stronger recent backbones such as VL-SAT and OCRL, but also for earlier backbones such as SGPN and SGFN, further supporting that the proposed supervision provides a backbone-agnostic structural signal.

The two downstream generators still differ in their visual style, shape priors, and layout realization, but they exhibit the same qualitative pattern: when the predicted graph is structurally closer to the reference graph, the generated scene better preserves the intended object set and spatial layout. This agreement mirrors the FID/KID improvements and further supports the generator-agnostic nature of our objective.

**Discussion.** The results support the central motivation of our method. A 3D scene graph is not merely a collection of independent object and predicate labels; it is a structured set of mutually dependent subject-predicate-object facts. When such a graph is used as a condition for 3D scene generation, local prediction errors can propagate into global errors in object composition, spatial arrangement, and relational plausibility. Therefore, optimizing only local classification losses may be insufficient, even when the resulting object or predicate predictions are individually reasonable.

Our graph-level supervision directly addresses this issue. The aligned triplet term $D^2_{\mathrm{align}}$ encourages each predicted subject-predicate-object fact to be close to its ground-truth counterpart in the contextual Hellinger triplet space, while the scene-internal geometry term $D^2_{\mathrm{geom}}$ encourages the predicted graph to preserve the relative distance geometry among relational facts within the scene. As a result, the predicted graph better captures both aligned semantic facts and their scene-internal relational geometry. The quantitative FID/KID improvements and qualitative reconstructions jointly demonstrate that this structured supervision improves the downstream usability of predicted 3D scene graphs, without requiring any change to the graph-conditioned scene generator.

### C.2 Qualitative Results on Generator-Limited Scene Domains

The generation-based FID/KID evaluation is restricted to scenes whose converted graphs are compatible with the 3D-FRONT-based graph-conditioned generators, as described in Appendix B.4. This filtering is applied using the reference graphs and is independent of the compared prediction methods; therefore, it does not introduce method-specific scene selection. However, it inevitably limits the room-domain coverage of the downstream generation evaluation, because several domain-specific objects in laundry, bathroom, or facility-heavy kitchen-like scenes cannot be represented by the current generator interface. To complement the generator-based evaluation, we additionally inspect the predicted scene graphs before graph-to-generator conversion.

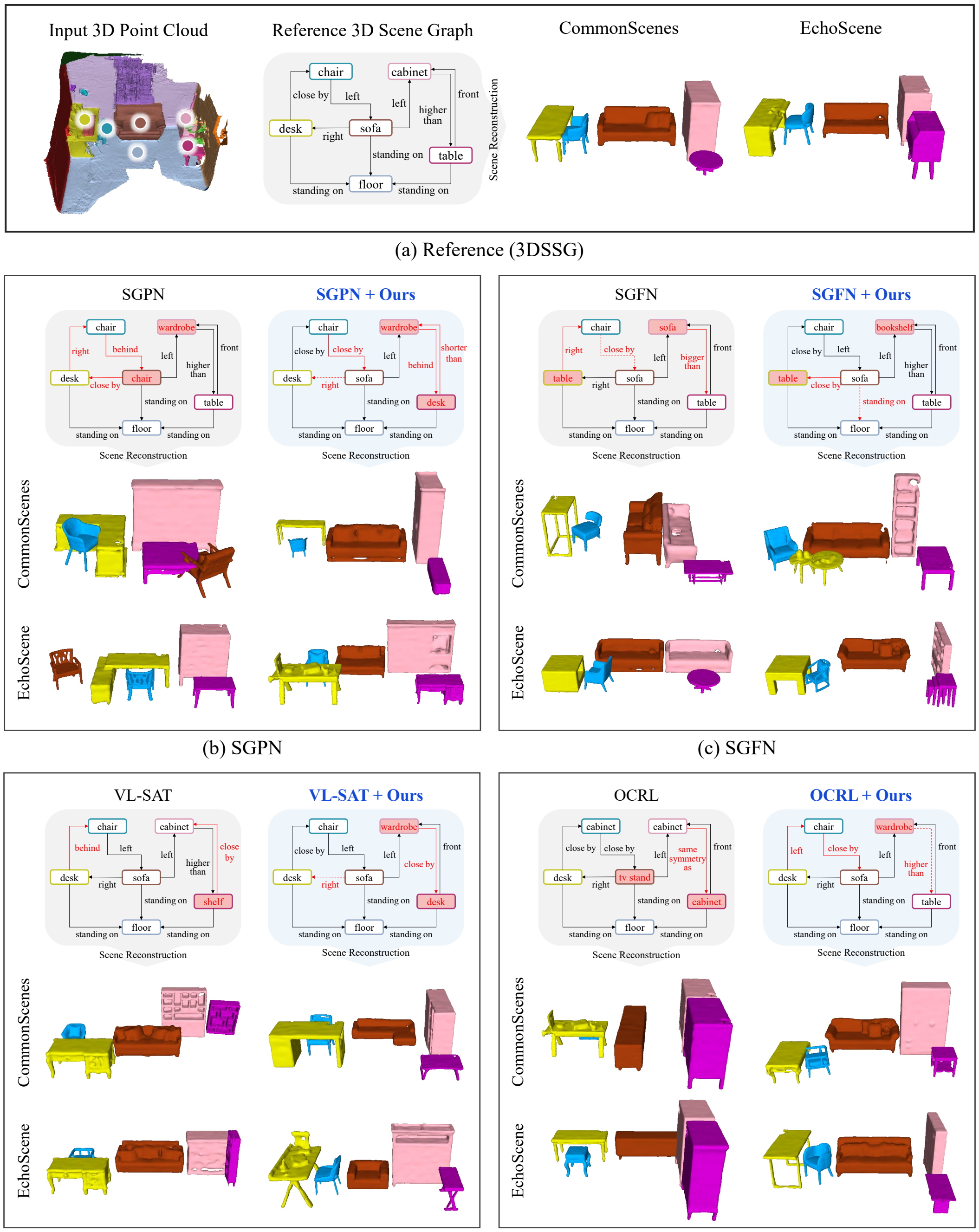

(a) Reference (3DSSG)

(b) SGPN

(c) SGFN

(d) VL-SAT

(e) OCRL

Figure 4: Qualitative comparison of graph-conditioned scene generation across different 3D scene graph prediction backbones and downstream generators. (a) Input 3D point cloud, reference 3DSSG graph, and scene reconstructions generated from the reference graph by CommonScenes [43] and EchoScene [44]. (b)–(e) For each backbone, we compare the original predictor on the left with the corresponding "+ Ours" variant on the right, followed by the scenes generated from each reconstructed graph by CommonScenes and EchoScene. Red annotations in the reconstructed graphs highlight predicted object or relation labels that differ from the reference graph, and blue titles indicate variants trained with our graph-level supervision. Our supervision yields more coherent graph reconstructions, leading to generated scenes that better preserve the reference object composition and spatial layout across both downstream generators.

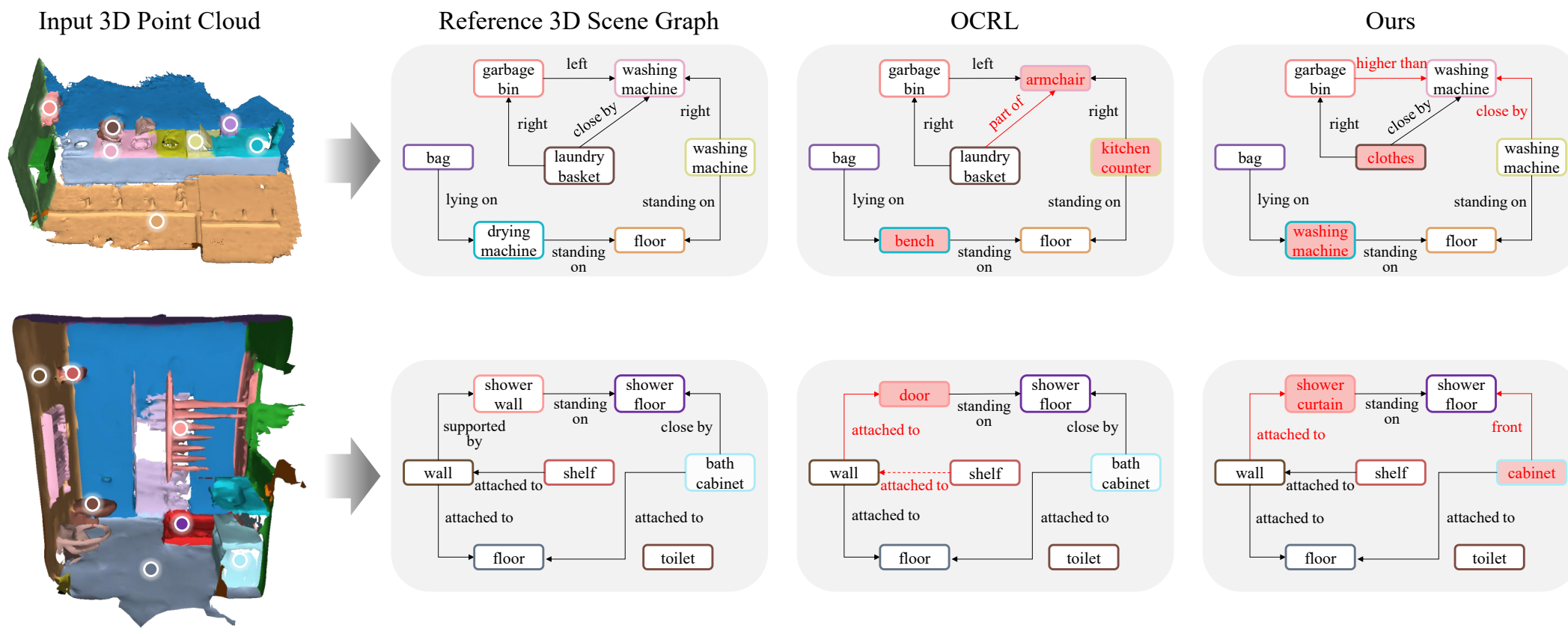

Figure 5: Qualitative graph prediction examples from generator-limited scene domains. Because the 3D-FRONT-based graph-conditioned generators support a restricted set of room types and object categories, laundry- and bathroom-like scenes are difficult to include in the generation-based FID/KID evaluation. We therefore compare the predicted graphs directly, before graph-to-generator conversion. OCRL often produces locally plausible but domain-inconsistent graphs, such as replacing laundry appliances with generic furniture or replacing shower-related components with unrelated objects. In contrast, our method better preserves domain-compatible object semantics and the relational scaffold of the scene.

Figure 5 shows qualitative graph prediction examples from scene domains that are difficult to evaluate through the current 3D-FRONT-based generation protocol. These examples highlight a failure mode that is not fully captured by local object or predicate accuracy: a predicted graph can remain locally connected and syntactically plausible while losing the functional identity of the scene.

In the laundry-like example, OCRL preserves several simple spatial relations such as *lying on* and *standing on*, but changes the semantic identity of the scene by replacing appliance-centric objects with unrelated furniture categories, such as *armchair*, *kitchen counter*, and *bench*. This type of error is particularly harmful for scene-graph-based reasoning, because the graph still contains objects and relations but no longer describes a coherent laundry scene. In contrast, our method better preserves the appliance-oriented semantics of the scene, including washing-machine-related objects and their support, proximity, and directional relations. Although the prediction is not identical to the reference graph, its errors remain more semantically compatible with the input scene domain.

A similar pattern appears in the bathroom-like example. OCRL keeps some generic structural elements such as *wall*, *floor*, *shelf*, and *toilet*, but replaces a shower-related component with *door*, which weakens the functional interpretation of the bathroom scene. Our method predicts a more domain-compatible shower-related object, *shower curtain*, and better maintains the attachment and support structure among the wall, shelf, floor, toilet, and cabinet-like furniture. This suggests that the proposed supervision does not merely improve isolated label predictions, but helps the graph preserve a coherent scene-level semantic scaffold.

These observations are consistent with the design of our objective. The aligned triplet term $D^2_{\text{align}}$ encourages each subject-predicate-object fact to remain close to its ground-truth counterpart in the contextual Hellinger triplet space, while the scene-internal geometry term $D^2_{\text{geom}}$ encourages the relative organization among relational facts to be preserved within the scene. As a result, the predicted graph is less likely to drift toward a set of individually plausible but jointly inconsistent object and relation labels. Importantly, these examples are evaluated directly at the graph level, without relying on a downstream generator. They therefore provide complementary evidence that the proposed graph-level supervision improves the predicted 3D scene graph itself, including in room domains that are under-represented or excluded in the generator-based quantitative evaluation.

Table 7: Conventional 3D scene graph prediction results on the 3DSSG [40] validation set. Following the standard evaluation protocol, we report exact-label recall (R) and mean recall (mR) for object, predicate, and triplet prediction. Higher values indicate better performance.

| Model | Object | | | | Predicate | | | | Triplet | | | |
|---|---|---|---|---|---|---|---|---|---|---|---|---|
| | R@1 | R@5 | mR@1 | mR@5 | R@1 | R@3 | mR@1 | mR@3 | R@50 | R@100 | mR@50 | mR@100 |
| SGPN [40] | 49.46 | 73.99 | 10.86 | 27.54 | 86.92 | 94.76 | 32.01 | 55.22 | 85.38 | 88.59 | 41.52 | 51.92 |
| SGFN [42] | 53.36 | 76.88 | 21.24 | 46.68 | 89.00 | 97.71 | 41.89 | 70.82 | 88.59 | 91.14 | 58.37 | 67.61 |
| VL-SAT [41] | 55.93 | 78.06 | 21.41 | 46.14 | 89.81 | 98.46 | 54.03 | 77.67 | 89.35 | 92.20 | 65.09 | 73.59 |
| OCRL [17] | **59.53** | 81.20 | 22.55 | 48.10 | **91.27** | 98.48 | 56.32 | 76.26 | 91.40 | 93.80 | 65.31 | 74.54 |
| Ours | 58.32 | **82.47** | **22.83** | **49.15** | 90.36 | **98.82** | **56.87** | **77.92** | **91.78** | **94.28** | **66.24** | **75.92** |

### C.3 Consistency with Conventional 3D Scene Graph Prediction Metrics

For completeness, we additionally evaluate our method under the conventional 3D scene graph prediction protocol used in prior 3D scene graph prediction studies [12, 17, 27–29, 40, 41]. Following this protocol, we report recall (R) [26] and mean recall (mR) [9] for object, predicate, and triplet prediction on the 3DSSG [40] validation set. Recall measures overall exact-label retrieval performance, while mean recall averages recall over classes and therefore reflects class-balanced performance under the long-tailed label distribution. For triplet prediction, a prediction is counted as correct only when the subject, predicate, and object labels all match the ground-truth triplet. The results are summarized in Table 7.

Our method remains highly competitive under this standard evaluation protocol. Among the compared methods, it achieves the best performance in 10 out of 12 reported metrics. For object prediction, our method improves R@5, mR@1, and mR@5 over the strongest competing results by 1.27, 0.28, and 1.05, respectively. For predicate prediction, it improves R@3 and mR@1 by 0.34 and 0.55, respectively, and achieves the best mR@3 with a 0.25 improvement over VL-SAT [41]. Most notably, our method achieves the best result on all triplet metrics, improving R@50, R@100, mR@50, and mR@100 over OCRL [17] by 0.38, 0.48, 0.93, and 1.38, respectively.

These results are particularly meaningful because our objective is not a direct local object and predicate classification loss. Prior methods are typically trained with node- and edge-level classification objectives, such as object cross-entropy and predicate cross-entropy or binary cross-entropy variants. In contrast, our method supervises the prediction through a graph-level distance between the predicted and ground-truth probabilistic scene graphs. The aligned triplet term $D^2_{\mathrm{align}}$ compares subject-predicate-object semantic facts in the contextual Hellinger triplet space, while the scene-internal geometry term $D^2_{\mathrm{geom}}$ encourages the predicted graph to preserve the relative distance geometry among relational facts. Despite not being explicitly optimized for independent top-1 object and predicate classification, our method achieves state-of-the-art-level conventional recall.

The only metrics where OCRL obtains higher scores are object R@1 and predicate R@1. This is expected because top-1 exact-label recall is the metric most directly aligned with local classification supervision. However, the gaps are small, and our method outperforms OCRL in higher-rank recall, mean recall, and all triplet metrics. The improvements in mean recall suggest that the proposed graph-level supervision does not simply favor frequent classes, but improves class-balanced prediction quality. The triplet improvements are especially important because triplet prediction better reflects whether object and predicate predictions are jointly consistent. This agrees with the design of our objective, which treats each directed object pair as a contextual subject-predicate-object fact rather than supervising object and predicate labels independently.

This supplementary comparison also clarifies the role of our main evaluation protocol. Conventional recall and mean recall are valuable for verifying exact-label prediction performance and for maintaining comparability with prior 3D scene graph prediction methods. However, they primarily measure local label agreement and do not directly evaluate whether the predicted graph preserves the semantic composition and relational organization needed to reconstruct a coherent 3D scene. For this reason, our main experiments evaluate predicted graphs through graph-conditioned scene generation, qualitative reconstruction, and user studies, as reported in Tables 1, 6, and 2. Together, these results show that the proposed graph-level supervision is compatible with standard 3D scene prediction metrics while providing stronger scene-level graph fidelity for downstream 3D scene generation.

### C.4 Ablation on the Candidate-Edge Setting

The main experiments use the all-pair edge setting, where the active edge-slot set is the complete directed object-pair set $\mathcal{A}$ defined in Eq. (2). This setting provides the most direct instantiation of our fixed-slot graph distance because every object pair can contribute to the contextual triplet field. In practical 3D scene graph systems, however, relation prediction is often performed over a restricted set of candidate object pairs for computational efficiency or architectural compatibility. We therefore further evaluate whether the proposed graph-level supervision remains effective when the active relation slots are restricted from $\mathcal{A}$ to a fixed candidate edge set $E^*$, as used in the candidate-edge objective in Eq. (37).

For a fair comparison, all variants in this ablation use the same OCRL backbone [17], the same EchoScene generator [44], the same 3DSSG train/validation split, and the same downstream graph-conditioned generation protocol as in Sec. 4.2. We also keep the training schedule, inference procedure, graph decoding rule, and rendering-based FID/KID evaluation unchanged. Thus, the comparison isolates the effect of the edge-slot setting and the corresponding supervision objective.

**Candidate-edge Construction.** We construct $E^*$ using only geometric information from object bounding boxes, without using ground-truth relation labels or model predictions. First, we select spatially nearby object pairs based on bounding-box center distance. For each object, we compute its center distance to the other objects in the scene and add the $K$ nearest neighboring objects as candidate pairs. This nearest-neighbor component is intended to cover relation candidates that are likely to arise from spatial proximity or relative layout, such as *close by*, *left*, *right*, *front*, and *behind*.

Second, we include an overlap-based component to better cover support or contact-related pairs that may not be selected by center distance alone. For each object, we compute its 3D bounding-box overlap with the other objects and add the $M$ objects with the largest overlap scores as additional candidates. This component is useful for pairs such as *chair–floor*, *table–floor*, or *shelf–wall*, where relations such as *standing on*, *lying on*, *attached to*, or *part of* may be important even when object centers are not the closest.

The final candidate edge set is the union of the center-distance-based and overlap-based candidates. Because 3D scene graph predicates are directed, each selected object pair is converted into bidirectional relation slots. We then remove self edges and duplicated directed pairs. In our experiments, we use $K = 3$ and $M = 2$. The resulting $E^*$ is fixed during both training and evaluation, and is independent of the ground-truth relation annotations and predicted relation scores.

**Compared Settings.** Table 8 compares three settings. The first row, *All-pair (Full)*, corresponds to the main objective in Eq. (36), where $D^2_{\mathrm{align}}$ and $D^2_{\mathrm{geom}}$ are computed over the complete directed edge-slot set $\mathcal{A}$. This is the main setting used in our experiments.

The second row, *Candidate-edge w/o* $D^2_{\mathrm{node}}$, replaces $\mathcal{A}$ with the fixed candidate set $E^*$ and computes the aligned triplet and geometry terms only over the active candidate relation slots. This variant tests whether the proposed graph-level supervision remains useful when relation coverage is restricted, but without explicitly adding a separate node-level metric block.

The third row, *Candidate-edge w/* $D^2_{\mathrm{node}}$, uses the candidate-edge objective in Eq. (37). That is, it combines $D^2_{\mathrm{align},E^*}$ and $D^2_{\mathrm{geom},E^*}$ with the node-augmented term $D^2_{\mathrm{node}}$. The node term is not a local cross-entropy loss; it is the squared Hellinger metric block between lifted object measures, as defined in Eq. (38). Its role is to preserve object-level semantic information when the active edge set is incomplete.

**Quantitative Results.** The all-pair full setting achieves the best generation quality, with an FID of $49.93$ and a KID of $2.78$. This result is expected because the all-pair formulation exposes the objective to the complete directed object-pair structure. Each object distribution participates repeatedly as a subject or object factor across many contextual triplets, and the geometry term can compare the relative arrangement of relational facts over the full edge-slot set. The all-pair setting therefore provides the richest supervision signal for learning scene-level graph structure.

When the supervision is restricted to the candidate edge set $E^*$, performance decreases to an FID of $52.48$ and a KID of $3.24$ without $D^2_{\mathrm{node}}$. This degradation reflects the reduced relational coverage of the candidate-edge setting. Although the nearest-neighbor and overlap-based rules capture many spatially local, support-related, and contact-related relations, they do not provide the same exhaustive coverage as $\mathcal{A}$. Consequently, some long-range or less geometrically obvious relations may be excluded from the active triplet field, and each object contributes to fewer contextual subject-predicate-object factors.

Table 8: Ablation on candidate-edge variants. We compare the all-pair graph-level objective with candidate-edge objectives with and without the node-augmented term $D^2_{\text{node}}$. All variants use the same OCRL backbone and EchoScene generator. KID scores are reported as $\times 10^{-3}$. Lower values indicate better generation quality.

| **Setting** | **Edge Set** | $D^2_{\text{align}}$ | $D^2_{\text{geom}}$ | $D^2_{\text{node}}$ | **FID** ↓ | **KID** ↓ |
|---|---|---|---|---|---|---|
| All-pair (Full) | $\mathcal{A}$ | ✓ | ✓ | – | **49.93** | **2.78** |
| Candidate-edge, w/o $D^2_{\text{node}}$ | $E^*$ | ✓ | ✓ | – | 52.48 | 3.24 |
| Candidate-edge, w/ $D^2_{\text{node}}$ | $E^*$ | ✓ | ✓ | ✓ | 51.03 | 2.98 |

Nevertheless, the candidate-edge graph-level objective still improves over the OCRL baseline in Table 1, which obtains an FID of $54.62$ and a KID of $3.76$ under the same downstream generator. This indicates that the proposed graph-level supervision remains beneficial even when it is applied only to a restricted candidate edge set. In other words, the gains of our method do not rely solely on exhaustive all-pair supervision; the aligned triplet and geometry terms still provide a useful structured signal under a practical candidate-edge regime.

Adding the node-augmented term further improves the candidate-edge setting. The FID decreases from $52.48$ to $51.03$, and the KID decreases from $3.24$ to $2.98$. Thus, $D^2_{\text{node}}$ improves the candidate-edge objective by $1.45$ FID points and $0.26$ KID points. Measured relative to the gap between the candidate-edge variant without $D^2_{\text{node}}$ and the all-pair full setting, the node-augmented term closes more than half of the degradation in both FID and KID. This supports the motivation of Eq. (37): when relation slots are sparse or incomplete, explicit node-level metric supervision helps preserve object semantics that may otherwise be only weakly represented through the selected triplet factors.

**Discussion.** These results show a clear trade-off between relational coverage and practical candidate-edge restriction. The all-pair objective remains the strongest setting because it compares predicted and ground-truth graphs over the complete directed object-pair structure. This is particularly important for our graph-level distance: $D^2_{\text{align}}$ aligns each contextual subject-predicate-object fact, while $D^2_{\text{geom}}$ preserves the scene-internal geometry among relational facts. Both terms become most informative when they are computed over the full relational field.

At the same time, the candidate-edge results demonstrate that our formulation is not limited to the all-pair setting. Even when the active edge set is restricted to $E^*$, the graph-level objective still improves downstream scene generation quality compared with local OCRL supervision. This suggests that the proposed supervision captures structure that is useful for graph-conditioned generation beyond independent object and predicate classification. The predicted graph becomes a better condition for the generator because the training objective compares relational facts as contextual triplets and regularizes their scene-internal arrangement.

The node-augmented term is especially important in this restricted setting. In the all-pair formulation, object semantics are repeatedly embedded into triplet measures through all directed object-pair slots. In contrast, under candidate-edge supervision, each object participates in only a subset of relation slots, and in degenerate cases an object may have very limited relational participation. The term $D^2_{\text{node}}$ compensates for this loss of coverage by directly preserving the lifted object distributions while remaining in the same Hellinger-based metric framework as the triplet terms. As a result, the candidate-edge variant with $D^2_{\text{node}}$ substantially reduces the gap to the all-pair full objective.

Overall, this ablation validates two design choices. First, using the all-pair edge set as the main setting is justified because it yields the best downstream generation quality and provides the most complete graph-level supervision. Second, the proposed framework can be adapted to candidate-edge systems by adding the node-augmented metric block. This makes the method compatible with practical relation-candidate pipelines while preserving its central advantage: training the predictor to produce scene graphs that are semantically and structurally useful as graph-conditioned generation inputs.

### C.5 Failure Cases and Limitations

Although the proposed graph-level supervision consistently improves scene-level graph fidelity, it is not free from failure cases. Figure 6 shows a representative example where our method preserves the overall semantic and relational structure of the scene, but fails to recover a contextually atypical object and some fine-grained directional relations. The input scene is a kitchen-like environment containing a counter, sink, wall, floor, and multiple cabinet-like structures. In the reference graph, however, one object is annotated as a *commode*, which is relatively unusual in this surrounding context. Our prediction preserves most of the global graph scaffold, including the *standing on*, *build in*, *attached to*, and *higher than* relations, but predicts the *commode* as a *kitchen cabinet*. It also relaxes the directional predicates *front* and *right* into the broader proximity predicate *close by*.

This example illustrates an important property of our method. The contextual Hellinger triplet representation encourages the predicted graph to be semantically coherent with the surrounding scene context and to preserve the scene-internal relational geometry. This is beneficial in most cases, as shown by the quantitative improvements in Tables 1 and 6, the qualitative results in Figures 2 and 4, and the user study in Table 2. However, when an object or relation is rare, ambiguous, or weakly supported by the surrounding context, the same contextual consistency can bias the prediction toward a more common scene-compatible alternative. In Figure 6, the predicted *kitchen cabinet* is not a random or implausible label; rather, it is a contextually plausible substitution given the nearby kitchen counter, sink, wall, and cabinet structures. The limitation is therefore not a collapse of the graph structure, but a loss of fidelity to an atypical fine-grained annotation.

A similar effect appears in relation prediction. Fine-grained directional predicates such as *front* and *right* require precise geometric disambiguation and can be difficult to distinguish in cluttered indoor layouts where several objects are spatially adjacent. In such cases, our graph-level objective may prefer a more stable proximity relation such as *close by*, which preserves the fact that the two objects are spatially related but loses the exact directionality. This behavior is consistent with the design of our objective: $D^2_{\mathrm{align}}$ encourages semantic agreement of aligned subject-predicate-object facts, while $D^2_{\mathrm{geom}}$ encourages the relative organization among relational facts to remain coherent. As a result, the predicted graph may remain useful for downstream scene reconstruction even when some exact object or predicate labels are softened.

These observations suggest two main limitations. First, because the contextual semantic measures are estimated from the training scene graph distribution, rare object-context combinations and long-tailed relation patterns can still be challenging. Second, relations that depend on subtle metric geometry or viewpoint-sensitive orientation may require stronger geometric cues than those captured by the current graph-level semantic objective alone. A promising direction is to combine the proposed graph-level supervision with rare-context calibration, uncertainty-aware multi-hypothesis prediction, or explicit geometric constraints for directional predicates. Such extensions could improve fine-grained label fidelity while retaining the main advantage of our method: producing semantically coherent and structurally useful scene graphs for downstream 3D scene generation.

### C.6 Broader Impacts

This work studies graph-level supervision for fixed-correspondence 3D scene graph prediction and evaluates predicted graphs through downstream graph-conditioned 3D scene generation. Potential positive impacts include improved structured 3D scene understanding for robotics, AR/VR, embodied AI, indoor scene simulation, and assistive spatial reasoning systems. By encouraging predicted scene graphs to preserve scene-level semantic composition and relational geometry, the proposed objective may help downstream systems use more coherent symbolic representations of 3D environments.

Potential negative impacts arise from the broader use of 3D scene understanding and graph-conditioned indoor scene generation. More accurate scene graph prediction may support privacy-sensitive mapping of indoor spaces, surveillance-oriented spatial analytics, or synthetic reconstruction of private environments if deployed without appropriate consent and access control. The method can also inherit biases from the underlying 3DSSG/3RScan data and from the restricted vocabularies of graph-conditioned generators, potentially underrepresenting uncommon room layouts, cultural variations in indoor spaces, rare object categories, or unusual object-relation configurations. In safety-critical embodied applications, plausible but incorrect predicted relations could also lead to inappropriate downstream planning decisions.

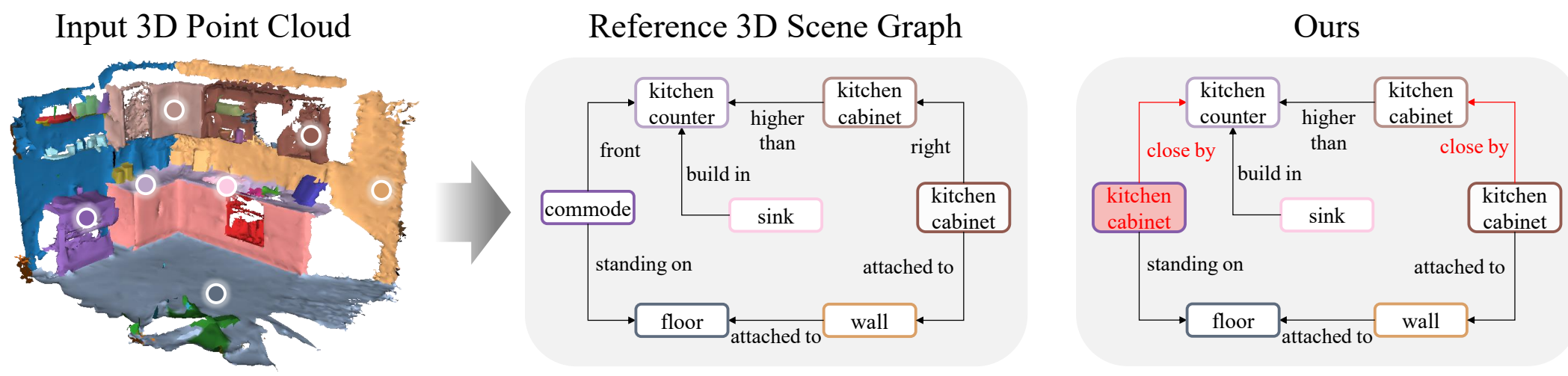

Figure 6: Failure case of the proposed graph-level supervision. The input point cloud depicts a kitchen-like scene, and the reference 3D scene graph contains a contextually atypical object label, *commode*, together with fine-grained directional predicates such as *front* and *right*. Our method preserves the overall relational scaffold of the scene, including *standing on*, *build in*, *attached to*, and *higher than*, but predicts the atypical *commode* as a contextually plausible *kitchen cabinet* and relaxes some directional predicates into the broader proximity relation *close by*. This case illustrates that the contextual semantic prior improves global graph coherence in general, but can occasionally favor common scene-compatible labels over rare or fine-grained annotations.

The present work is not a deployed system and does not release raw indoor scans, private user data, or a high-risk generative model. Mitigations for downstream use include obtaining consent for real-world indoor scans, avoiding deployment in surveillance or access-control settings without separate ethical review, auditing performance across scene domains and object categories, and keeping human oversight when predicted scene graphs are used by embodied agents or generative systems.

# D Full Construction of Contextual Semantic Measures

This appendix specifies the count-based construction underlying the contextual object measures in Eq. (15) and the contextual relation measures in Eq. (16). It also records the canonical no-relation measure in Eq. (17), which is not estimated from empirical counts.

All data-driven statistics are computed once from the training scene graph database and are kept fixed during model training. The no-relation row is set analytically by Eq. (17).

Let $\mathcal{D}_{\text{train}}$ denote the training scene graph database. For each training scene $S \in \mathcal{D}_{\text{train}}$, let $V_S$ be its object-slot set, let $y_i^S \in \mathcal{C}$ be the ground-truth object label of node slot $i \in V_S$, and let $Y_{(i,j)}^S \subseteq \mathcal{R}$ be the ground-truth positive predicate set of the directed object-pair slot $(i, j)$. A *positive relation occurrence* is a tuple $(S, i, r, j)$ such that $r \in Y_{(i,j)}^S$.

## D.1 Object Context Space and Object Measures

The object context space contains identity coordinates and positive incoming and outgoing relational-context coordinates:

$$\Omega_V = \{\text{id} : c \mid c \in \mathcal{C}\} \cup \{\text{out} : r, d \mid r \in \mathcal{R},\ d \in \mathcal{C}\} \cup \{\text{in} : d, r \mid d \in \mathcal{C},\ r \in \mathcal{R}\}. \tag{44}$$

For each object class $c \in \mathcal{C}$, we maintain a count table

$$N_V(c, \omega), \qquad \omega \in \Omega_V. \tag{45}$$

The table is initialized to zero and updated as follows.

For every node occurrence with ground-truth label $y_i^S = c$, we increment the class-specific identity coordinate:

$$N_V(c, \text{id} : c) \leftarrow N_V(c, \text{id} : c) + 1. \tag{46}$$

For every positive relation occurrence $(S, i, r, j)$ with $y_i^S = c_s$ and $y_j^S = c_o$, we increment the outgoing context of the subject class and the incoming context of the object class:

$$N_V(c_s, \text{out} : r, c_o) \leftarrow N_V(c_s, \text{out} : r, c_o) + 1, \tag{47}$$

$$N_V(c_o, \text{in} : c_s, r) \leftarrow N_V(c_o, \text{in} : c_s, r) + 1. \tag{48}$$

Only positive relations are used in Eqs. (47) and (48). No-relation pairs are excluded from the object context construction. This prevents the contextual object measures from being dominated by the large number of absent directed object pairs in all-pair formulations.

After optional smoothing described in Appendix D.5, the normalized object measure is

$$\mu_V(c)(\omega) = \frac{\widetilde{N}_V(c,\omega)}{\sum_{\omega' \in \Omega_V} \widetilde{N}_V(c,\omega')}, \qquad \omega \in \Omega_V. \tag{49}$$

Equation (49) instantiates the contextual object measure referenced in Eq. (15).

### D.2 Relation Context Space and Relation Measures

The relation context space contains predicate identity coordinates together with subject, object, and subject-object pair context coordinates:

$$\Omega_R = \{\text{id} : r \mid r \in \mathcal{R}_0\} \cup \{\text{subj} : c \mid c \in \mathcal{C}\} \cup \{\text{obj} : c \mid c \in \mathcal{C}\} \cup \{\text{pair} : c_s, c_o \mid c_s, c_o \in \mathcal{C}\}. \tag{50}$$

For each positive predicate label $r \in \mathcal{R}$, we maintain a count table

$$N_R(r, \omega), \qquad \omega \in \Omega_R. \tag{51}$$

The table is initialized to zero and updated from positive relation occurrences.

For every positive relation occurrence $(S, i, r, j)$ with $y_i^S = c_s$ and $y_j^S = c_o$, we increment the following entries:

$$N_R(r, \text{id} : r) \leftarrow N_R(r, \text{id} : r) + 1, \tag{52}$$

$$N_R(r, \text{subj} : c_s) \leftarrow N_R(r, \text{subj} : c_s) + 1, \tag{53}$$

$$N_R(r, \text{obj} : c_o) \leftarrow N_R(r, \text{obj} : c_o) + 1, \tag{54}$$

$$N_R(r, \text{pair} : c_s, c_o) \leftarrow N_R(r, \text{pair} : c_s, c_o) + 1. \tag{55}$$

After optional smoothing described in Appendix D.5, the normalized relation measure for a positive predicate $r \in \mathcal{R}$ is

$$\mu_R(r)(\omega) = \frac{\widetilde{N}_R(r,\omega)}{\sum_{\omega' \in \Omega_R} \widetilde{N}_R(r,\omega')}, \qquad \omega \in \Omega_R. \tag{56}$$

Equation (56) instantiates the contextual relation measure referenced in Eq. (16) for all positive predicate classes.

### D.3 Identity Anchors and Injective Signatures

The identity coordinates in Eqs. (44) and (50) are the key structural device that makes the contextual lifting injective in Appendix F.2. Throughout this subsection, $\mathcal{C}$ and $\mathcal{R}$ are understood as the effective vocabularies obtained after removing labels that never appear in $\mathcal{D}_{\text{train}}$. For object classes, the count construction enforces

$$N_V(c, \text{id} : c) > 0 \quad \text{whenever class } c \text{ appears in } \mathcal{D}_{\text{train}}, \tag{57}$$

and, for $c' \neq c$,

$$N_V(c', \text{id} : c) = 0. \tag{58}$$

Because the smoothing rule in Appendix D.5 never adds mass to foreign object identity coordinates, Eqs. (57) and (58) imply

$$\mu_V(c)(\text{id} : c) > 0, \qquad \mu_V(c')(\text{id} : c) = 0 \quad \text{for } c' \neq c. \tag{59}$$

Similarly, for each positive predicate class $r \in \mathcal{R}$ that appears in the training database,

$$N_R(r, \mathrm{id} : r) > 0, \tag{60}$$

and, for $r' \in \mathcal{R} \setminus \{r\}$,

$$N_R(r', \mathrm{id} : r) = 0. \tag{61}$$

Together with the canonical no-relation row in Eq. (17), the smoothing rule preserves these support constraints, so for every $r \in \mathcal{R}$,

$$\mu_R(r)(\mathrm{id} : r) > 0, \qquad \mu_R(r')(\mathrm{id} : r) = 0 \quad \text{for } r' \in \mathcal{R}_0 \setminus \{r\}. \tag{62}$$

These anchor-support properties are exactly the ingredients needed in Appendix F.2 to recover the original object distributions from the lifted measures in Eq. (18) and the positive predicate components of the lifted relation measures in Eq. (20). Recovery of the no-relation component is additionally provided by the dedicated coordinate in Eq. (17).

### D.4 Canonical No-Relation Measure

As stated in Eq. (17), the no-relation predicate is represented by the canonical absence measure.

This row is not estimated from empirical negative-pair counts. The reason is that, in all-pair formulations, absent directed object pairs are overwhelmingly more frequent than positive predicate occurrences. Estimating the no-relation row from empirical negative statistics would therefore entangle the semantic meaning of no-relation with dataset-level object co-occurrence priors and could dominate the geometry of the contextual predicate space.

By using the canonical measure in Eq. (17), no-relation acts only as an explicit absence symbol. It retains its own predicate identity anchor $\mathrm{id} : \varnothing$, remains separated from positive predicates, and preserves the support condition required by the injectivity argument in Appendix F.2.

### D.5 Normalization and Rare-Label Smoothing

The contextual measures in Eqs. (49) and (56) may be smoothed to reduce instability for rare classes. However, smoothing must preserve the identity anchor structure in Eqs. (59) and (62). We therefore smooth only non-identity context coordinates.

Define the object identity token subset

$$\Omega_V^{\mathrm{id}} = \{\mathrm{id} : c \mid c \in \mathcal{C}\}, \tag{63}$$

and the relation identity token subset

$$\Omega_R^{\mathrm{id}} = \{\mathrm{id} : r \mid r \in \mathcal{R}_0\}. \tag{64}$$

Let $\varepsilon_V \geq 0$ and $\varepsilon_R \geq 0$ be smoothing constants. For object rows, we define

$$\widetilde{N}_V(c, \omega) = \begin{cases} N_V(c, \omega), & \omega \in \Omega_V^{\mathrm{id}}, \\ N_V(c, \omega) + \varepsilon_V, & \omega \in \Omega_V \setminus \Omega_V^{\mathrm{id}}. \end{cases} \tag{65}$$

For positive predicate rows, we define

$$\widetilde{N}_R(r, \omega) = \begin{cases} N_R(r, \omega), & \omega \in \Omega_R^{\mathrm{id}}, \\ N_R(r, \omega) + \varepsilon_R, & \omega \in \Omega_R \setminus \Omega_R^{\mathrm{id}}, \end{cases} \qquad r \in \mathcal{R}. \tag{66}$$

The no-relation row is not smoothed, since it is fixed by Eq. (17).

The contextual measures are then obtained by row normalization using Eqs. (49) and (56). When $\varepsilon_V = \varepsilon_R = 0$, this reduces to pure empirical normalization. When smoothing is used, the identity anchors remain untouched, so the support properties in Eqs. (59) and (62) continue to hold. The effective-vocabulary convention above is the vocabulary condition used in Theorem 3.1.

## E Hellinger Lifting and Product Measure Factorization

This appendix provides the technical details underlying the Hellinger lifting and triplet product representation used in Section 3. Throughout, we use the contextual object and relation measures from Eqs. (15) and (16), the lifted object and relation measures from Eqs. (18), (19), (20), and (21), and the Hellinger definitions in Eqs. (22) and (23).

### E.1 Lifted Predictions as Probability Measures

We first verify that the lifted predictions used in Section 3.1 are valid probability measures.

**Proposition E.1.** *For every node slot $i$ and every edge slot $a$, the lifted measures $\alpha_i^{pred}$, $\alpha_i^{gt}$, $\beta_a^{pred}$, and $\beta_a^{gt}$ in Eqs.* (18), (19), (20)*, and* (21) *are probability measures on $\Omega_V$ and $\Omega_R$, respectively.*

*Proof.* We begin with the predicted object lift in Eq. (18). Since $p_i \in \Delta^{|\mathcal{C}|}$ by Eq. (4), we have $p_i(c) \geq 0$ for all $c \in \mathcal{C}$ and

$$\sum_{c \in \mathcal{C}} p_i(c) = 1. \tag{67}$$

Each contextual object row $\mu_V(c)$ is a probability measure on $\Omega_V$ by Eq. (15), so

$$\mu_V(c)(\omega) \geq 0 \quad \text{for all } \omega \in \Omega_V, \qquad \sum_{\omega \in \Omega_V} \mu_V(c)(\omega) = 1. \tag{68}$$

Therefore, for every $\omega \in \Omega_V$,

$$\alpha_i^{pred}(\omega) = \sum_{c \in \mathcal{C}} p_i(c)\mu_V(c)(\omega) \geq 0, \tag{69}$$

and

$$\begin{aligned}\sum_{\omega \in \Omega_V} \alpha_i^{pred}(\omega) &= \sum_{\omega \in \Omega_V} \sum_{c \in \mathcal{C}} p_i(c)\mu_V(c)(\omega) \\ &= \sum_{c \in \mathcal{C}} p_i(c) \sum_{\omega \in \Omega_V} \mu_V(c)(\omega) \\ &= \sum_{c \in \mathcal{C}} p_i(c) = 1.\end{aligned} \tag{70}$$

Hence $\alpha_i^{pred}$ is a probability measure on $\Omega_V$.

The ground-truth object lift in Eq. (19) is simply a single contextual object row. Since each row $\mu_V(c)$ is a probability measure by Eq. (15), we have $\alpha_i^{gt} \in \Delta^{|\Omega_V|}$.

The same argument applies to the predicted relation lift in Eq. (20). Since $q_a \in \Delta^{|\mathcal{R}_0|}$ by Eq. (5),

$$\sum_{r \in \mathcal{R}_0} q_a(r) = 1, \qquad q_a(r) \geq 0 \text{ for all } r \in \mathcal{R}_0. \tag{71}$$

Each contextual predicate row $\mu_R(r)$ is a probability measure on $\Omega_R$ by Eq. (16), hence

$$\sum_{\omega \in \Omega_R} \beta_a^{pred}(\omega) = 1, \qquad \beta_a^{pred}(\omega) \geq 0 \text{ for all } \omega \in \Omega_R. \tag{72}$$

Thus $\beta_a^{pred}$ is a probability measure on $\Omega_R$.

Finally, consider the ground-truth relation lift in Eq. (21). If $Y_a^{gt} = \emptyset$, then $\beta_a^{gt} = \mu_R(\varnothing)$, which is a probability measure by Eq. (17). If $Y_a^{gt} \neq \emptyset$, then $\beta_a^{gt}$ is the uniform average of the probability measures $\{\mu_R(r)\}_{r \in Y_a^{gt}}$, and therefore

$$\sum_{\omega \in \Omega_R} \beta_a^{gt}(\omega) = \frac{1}{|Y_a^{gt}|} \sum_{r \in Y_a^{gt}} \sum_{\omega \in \Omega_R} \mu_R(r)(\omega) = 1, \tag{73}$$

with nonnegative coordinates. Hence $\beta_a^{gt}$ is also a probability measure. □

The same reasoning applies to the relation distributions obtained from sigmoid multi-label scores after the conversion described in Appendix G.3. Once that conversion yields a vector in $\Delta^{|\mathcal{R}_0|}$, Eq. (20) again produces a valid probability measure.

### E.2 Mixture Before Square-Root Embedding

A central design choice of the method is that the square root is applied after forming the semantic mixture, as stated in Eq. (24). This appendix explains why that order is essential.

**Proposition E.2.** *The features in Eq.* (24) *are the exact Hellinger embeddings of the lifted probability measures in Eqs.* (18)*,* (19)*,* (20)*, and* (21)*.*

*Proof.* By Proposition E.1, each lifted measure is a valid probability measure. The Hellinger embedding of a probability measure $P$ is its pointwise square root $\sqrt{P}$, and the corresponding Bhattacharyya coefficient is defined in Eq. (22). Equation (24) applies this embedding to each lifted object and relation measure. Thus $v_i^{pred}$, $v_i^{gt}$, $u_a^{pred}$, and $u_a^{gt}$ are exactly the square-root embeddings of the lifted measures. □

This point matters because the feature induced by Eq. (18) is

$$v_i^{pred}(\omega) = \sqrt{\sum_{c\in\mathcal{C}} p_i(c)\mu_V(c)(\omega)}, \qquad \omega \in \Omega_V, \tag{74}$$

and, analogously, the feature induced by Eq. (20) is

$$u_a^{pred}(\omega) = \sqrt{\sum_{r\in\mathcal{R}_0} q_a(r)\mu_R(r)(\omega)}, \qquad \omega \in \Omega_R. \tag{75}$$

In general,

$$\sqrt{\sum_k \lambda_k P_k(\omega)} \neq \sum_k \lambda_k \sqrt{P_k(\omega)}, \tag{76}$$

except in special cases, such as a single active component or identical component measures. Therefore, replacing the proposed construction with a weighted sum of pre-embedded class prototypes would not be the Hellinger embedding of the lifted probability measure. The proposed order in Eqs. (18), (20), and (24) is precisely what makes the method an exact Hellinger comparison between contextual semantic mixtures.

The node-level affinity induced by this construction is

$$\mathrm{BC}\left(\alpha_i^{pred}, \alpha_i^{gt}\right) = \left\langle v_i^{pred}, v_i^{gt}\right\rangle, \tag{77}$$

and the corresponding squared Hellinger distance is

$$H^2\left(\alpha_i^{pred}, \alpha_i^{gt}\right) = 1 - \left\langle v_i^{pred}, v_i^{gt}\right\rangle. \tag{78}$$

Similarly, at the predicate level,

$$\mathrm{BC}\left(\beta_a^{pred}, \beta_a^{gt}\right) = \left\langle u_a^{pred}, u_a^{gt}\right\rangle, \tag{79}$$

and

$$H^2\left(\beta_a^{pred}, \beta_a^{gt}\right) = 1 - \left\langle u_a^{pred}, u_a^{gt}\right\rangle. \tag{80}$$

### E.3 Triplet Product Measure Factorization

We now derive the triplet product measure representation used in Section 3.2. Let

$$\Xi = \Omega_V \times \Omega_R \times \Omega_V \tag{81}$$

denote the triplet context space. For any aligned probabilistic scene graph $G$, Eq. (25) instantiates a triplet measure $\rho_a^G \in \Delta^{|\Xi|}$ for each edge slot $a \in \mathcal{A}$. On the product space $\Xi$, this means

$$\rho_a^G(x,y,z) = \alpha_{s(a)}^G(x)\beta_a^G(y)\alpha_{t(a)}^G(z), \qquad (x,y,z)\in\Xi. \tag{82}$$

**Proposition E.3.** *For every aligned probabilistic scene graph $G$ and every edge slot $a$, the triplet product measure instantiated from Eq.* (25) *is a probability measure on $\Xi$.*

*Proof.* Nonnegativity follows immediately from the nonnegativity of the lifted measures. Moreover,

$$\begin{aligned}\sum_{(x,y,z)\in\Xi} \rho_a^G(x,y,z) &= \sum_{x\in\Omega_V}\sum_{y\in\Omega_R}\sum_{z\in\Omega_V} \alpha_{s(a)}^G(x)\beta_a^G(y)\alpha_{t(a)}^G(z) \\ &= \left(\sum_{x\in\Omega_V}\alpha_{s(a)}^G(x)\right)\left(\sum_{y\in\Omega_R}\beta_a^G(y)\right)\left(\sum_{z\in\Omega_V}\alpha_{t(a)}^G(z)\right) \\ &= 1. \end{aligned} \tag{83}$$

Hence $\rho_a^G$ is a probability measure on $\Xi$. ☐

**Proposition E.4.** *The square-root Hellinger feature of the triplet product measure $\rho_a^G$ factorizes exactly as in Eq.* (26)*.*

*Proof.* Applying the pointwise square root to Eq. (82) gives

$$\begin{aligned}\sqrt{\rho_a^G(x,y,z)} &= \sqrt{\alpha_{s(a)}^G(x)}\sqrt{\beta_a^G(y)}\sqrt{\alpha_{t(a)}^G(z)} \\ &= v_{s(a)}^G(x)\, u_a^G(y)\, v_{t(a)}^G(z), \end{aligned} \tag{84}$$

where the second equality uses Eq. (24). In tensor product notation, Eq. (84) is exactly Eq. (26) instantiated for graph $G$. ☐

Equation (26) shows that each directed object pair is represented as a single contextual semantic fact. This is the mathematical reason the aligned term in Eq. (28) compares subject-predicate-object triplets rather than independent object and predicate labels.

## E.4 Bhattacharyya Affinity Factorization

We now derive the factorized triplet affinity used in Eq. (27). Consider two aligned probabilistic scene graphs $G$ and $H$, and two edge slots $a, b \in \mathcal{A}$. Using Eq. (26), the Bhattacharyya coefficient between the two triplet measures is

$$\begin{aligned}\mathrm{BC}\left(\rho_a^G, \rho_b^H\right) &= \left\langle \sqrt{\rho_a^G}, \sqrt{\rho_b^H} \right\rangle \\ &= \left\langle v_{s(a)}^G \otimes u_a^G \otimes v_{t(a)}^G,\, v_{s(b)}^H \otimes u_b^H \otimes v_{t(b)}^H \right\rangle. \end{aligned} \tag{85}$$

For tensor-product Hilbert spaces, the inner product factorizes:

$$\langle x_1 \otimes y_1 \otimes z_1,\, x_2 \otimes y_2 \otimes z_2 \rangle = \langle x_1, x_2\rangle \langle y_1, y_2\rangle \langle z_1, z_2\rangle. \tag{86}$$

Applying Eq. (86) to Eq. (85) yields

$$\mathrm{BC}\left(\rho_a^G, \rho_b^H\right) = \left\langle v_{s(a)}^G, v_{s(b)}^H \right\rangle \left\langle u_a^G, u_b^H \right\rangle \left\langle v_{t(a)}^G, v_{t(b)}^H \right\rangle. \tag{87}$$

Since the triplet Hellinger feature is the square-root embedding, the left-hand side of Eq. (87) is also

$$\mathrm{BC}\left(\rho_a^G, \rho_b^H\right) = \left\langle z_a^G, z_b^H \right\rangle. \tag{88}$$

Combining Eqs. (87) and (88), we recover the factorized triplet affinity in Eq. (27).

The aligned triplet term in Eq. (28) is the special case $b = a$, where

$$\left\langle z_a^G, z_a^H \right\rangle = \left\langle v_{s(a)}^G, v_{s(a)}^H \right\rangle \left\langle u_a^G, u_a^H \right\rangle \left\langle v_{t(a)}^G, v_{t(a)}^H \right\rangle. \tag{89}$$

Similarly, the within-graph Gram entry in Eq. (29) is the case $G = H$. Combining Eq. (87) with Eq. (29) recovers the factorized Gram entry implied by Eq. (27).

Finally, the Hellinger distance between triplet product measures is obtained by combining Eq. (23) with Eq. (88):

$$H^2\left(\rho_a^G, \rho_b^H\right) = 1 - \left\langle z_a^G, z_b^H \right\rangle. \tag{90}$$

For the within-graph case $G = H$, Eq. (90) specializes to the identity used in Eq. (31), which in turn yields the distance-geometry equivalence in Eqs. (32) and (33).

## F Metric Proof

This appendix proves the metric statement summarized in Section 3.4. We work under the all-pair setting of Eq. (2), with fixed node and edge correspondence and $n \geq 2$.

### F.1 Space of Aligned Probabilistic Scene Graphs

We consider the space of aligned probabilistic 3D scene graphs

$$\mathcal{G}_{\text{align}} = \prod_{i \in V} \Delta^{|\mathcal{C}|} \times \prod_{a \in \mathcal{A}} \Delta^{|\mathcal{R}_0|}, \tag{91}$$

where the node slot set $V$ and the directed edge-slot set $\mathcal{A}$ are defined in Eqs. (1) and (2). An element

$$G = \Big(\{p_i^G\}_{i \in V}, \{q_a^G\}_{a \in \mathcal{A}}\Big) \in \mathcal{G}_{\text{align}} \tag{92}$$

assigns an object distribution to every node slot and a predicate distribution to every directed object-pair slot under the fixed correspondence.

For any $G \in \mathcal{G}_{\text{align}}$, its lifted object and predicate measures are obtained by applying the same lifting rules as Eqs. (18) and (20):

$$\alpha_i^G := \sum_{c \in \mathcal{C}} p_i^G(c) \mu_V(c), \qquad i \in V, \tag{93}$$

$$\beta_a^G := \sum_{r \in \mathcal{R}_0} q_a^G(r) \mu_R(r), \qquad a \in \mathcal{A}. \tag{94}$$

The corresponding triplet product measure is given by Eq. (25), and its Hellinger feature is given by Eq. (26). By Proposition E.3, each triplet product measure $\rho_a^G$ is a probability measure on the triplet context space, so all Hellinger distances used below are well defined.

Throughout this appendix, graph identity is defined under the fixed correspondence:

$$G = H \quad \Longleftrightarrow \quad p_i^G = p_i^H \text{ for all } i \in V, \quad q_a^G = q_a^H \text{ for all } a \in \mathcal{A}. \tag{95}$$

The training annotation in Eq. (13) is label-valued for nodes and set-valued for predicates. When this annotation is used as a target for the metric objective, we identify it with canonical probability vectors. For node slot $i$, define

$$p_i^{gt}(c) = \mathbf{1}[y_i^{gt} = c], \qquad c \in \mathcal{C}. \tag{96}$$

For edge slot $a$, define

$$q_a^{gt}(r) = \begin{cases} 1, & Y_a^{gt} = \emptyset \text{ and } r = \varnothing, \\ \frac{1}{|Y_a^{gt}|}, & Y_a^{gt} \neq \emptyset \text{ and } r \in Y_a^{gt}, \\ 0, & \text{otherwise}, \end{cases} \qquad r \in \mathcal{R}_0. \tag{97}$$

Under these definitions, Eqs. (19) and (21) are exactly the contextual lifts of $p_i^{gt}$ and $q_a^{gt}$. Thus, the supervised target is embedded as a point in the same probabilistic graph space used in the metric proof below.

For convenience, we also introduce the unsquared versions of the aligned and geometry terms:

$$D_{\text{align}}(G, H) := \sqrt{D^2_{\text{align}}(G, H)}, \tag{98}$$

$$D_{\text{geom}}(G, H) := \sqrt{D^2_{\text{geom}}(G, H)}. \tag{99}$$

With Eqs. (98) and (99), the canonical distance in Eq. (35) may be viewed as the Euclidean norm of the two blocks $D_{\text{align}}$ and $D_{\text{geom}}$.

### F.2 Injectivity of Contextual Semantic Lifting

The metric proof relies on the fact that the contextual semantic lifting is injective. This is exactly why the identity-anchor construction in Appendix D.3 is required.

**Proposition F.1.** *The object lifting map*

$$T_V : \Delta^{|\mathcal{C}|} \to \Delta^{|\Omega_V|}, \qquad T_V(p) = \sum_{c\in\mathcal{C}} p(c)\mu_V(c), \tag{100}$$

*is injective.*

*Proof.* Fix a class $c \in \mathcal{C}$. By the object identity-anchor support property in Eq. (59), the coordinate $\mathrm{id} : c$ has positive mass only in the row $\mu_V(c)$ and zero mass in all other object rows. Therefore, the $\mathrm{id} : c$ coordinate of $T_V(p)$ satisfies

$$\begin{aligned} T_V(p)(\mathrm{id} : c) &= \sum_{c'\in\mathcal{C}} p(c')\mu_V(c')(\mathrm{id} : c) \\ &= p(c)\mu_V(c)(\mathrm{id} : c). \end{aligned} \tag{101}$$

Hence

$$p(c) = \frac{T_V(p)(\mathrm{id} : c)}{\mu_V(c)(\mathrm{id} : c)}. \tag{102}$$

Since Eq. (102) recovers every coordinate $p(c)$ from $T_V(p)$, the map $T_V$ is injective. □

**Proposition F.2.** *The predicate lifting map*

$$T_R : \Delta^{|\mathcal{R}_0|} \to \Delta^{|\Omega_R|}, \qquad T_R(q) = \sum_{r\in\mathcal{R}_0} q(r)\mu_R(r), \tag{103}$$

*is injective.*

*Proof.* For any positive predicate $r \in \mathcal{R}$, the relation identity-anchor support property in Eq. (62) implies that the coordinate $\mathrm{id} : r$ has positive mass only in the row $\mu_R(r)$ and zero mass in every other predicate row. Therefore,

$$\begin{aligned} T_R(q)(\mathrm{id} : r) &= \sum_{r'\in\mathcal{R}_0} q(r')\mu_R(r')(\mathrm{id} : r) \\ &= q(r)\mu_R(r)(\mathrm{id} : r), \end{aligned} \tag{104}$$

and hence

$$q(r) = \frac{T_R(q)(\mathrm{id} : r)}{\mu_R(r)(\mathrm{id} : r)}, \qquad r \in \mathcal{R}. \tag{105}$$

For the no-relation component, the count construction in Eqs. (52)–(55) and the smoothing rule in Eq. (66) imply that every positive predicate row satisfies

$$\mu_R(r)(\mathrm{id} : \varnothing) = 0, \qquad r \in \mathcal{R}. \tag{106}$$

Moreover, Eq. (17) gives

$$\mu_R(\varnothing)(\mathrm{id} : \varnothing) = 1. \tag{107}$$

Therefore,

$$\begin{aligned} T_R(q)(\mathrm{id} : \varnothing) &= \sum_{r\in\mathcal{R}} q(r)\mu_R(r)(\mathrm{id} : \varnothing) + q(\varnothing)\mu_R(\varnothing)(\mathrm{id} : \varnothing) \\ &= q(\varnothing). \end{aligned} \tag{108}$$

Equations (105) and (108) recover every coordinate of $q$ from $T_R(q)$, proving injectivity. □

**Corollary F.3.** *For two aligned probabilistic scene graphs $G, H \in \mathcal{G}_{\mathrm{align}}$,*

$$\alpha_i^G = \alpha_i^H \quad \Longrightarrow \quad p_i^G = p_i^H, \tag{109}$$

*and*

$$\beta_a^G = \beta_a^H \quad \Longrightarrow \quad q_a^G = q_a^H. \tag{110}$$

*Proof.* Apply Propositions F.1 and F.2 to the lifted measures in Eqs. (93) and (94). □

### F.3 Non-Negativity and Symmetry

We next verify the basic metric axioms.

**Proposition F.4.** *The functions $D_{\text{align}}$, $D_{\text{geom}}$, and $D_{\text{metric}}$ are nonnegative and symmetric on $\mathcal{G}_{\text{align}}$.*

*Proof.* By Eq. (28), $D^2_{\text{align}}(G,H)$ is the average of squared Hellinger distances between corresponding triplet product measures. Since $H^2(\cdot,\cdot) \geq 0$ by Eq. (23), we have

$$D^2_{\text{align}}(G,H) \geq 0, \qquad D_{\text{align}}(G,H) \geq 0. \tag{111}$$

The Hellinger distance is symmetric, so

$$D^2_{\text{align}}(G,H) = D^2_{\text{align}}(H,G), \qquad D_{\text{align}}(G,H) = D_{\text{align}}(H,G). \tag{112}$$

By Eq. (34), $D^2_{\text{geom}}(G,H)$ is an average of squared real differences, hence

$$D^2_{\text{geom}}(G,H) \geq 0, \qquad D_{\text{geom}}(G,H) \geq 0. \tag{113}$$

Using the triplet distance-geometry matrix from Eq. (30),

$$\left(\Gamma_G[a,b] - \Gamma_H[a,b]\right)^2 = \left(\Gamma_H[a,b] - \Gamma_G[a,b]\right)^2, \tag{114}$$

so Eq. (34) also gives

$$D^2_{\text{geom}}(G,H) = D^2_{\text{geom}}(H,G), \qquad D_{\text{geom}}(G,H) = D_{\text{geom}}(H,G). \tag{115}$$

Finally, Eq. (35) expresses $D_{\text{metric}}$ as the square root of a sum of two nonnegative symmetric terms, so

$$D_{\text{metric}}(G,H) \geq 0, \qquad D_{\text{metric}}(G,H) = D_{\text{metric}}(H,G). \tag{116}$$

□

### F.4 Triangle Inequality

We now prove the triangle inequality. The proof is clearest when the aligned and geometry terms are treated separately.

**Proposition F.5.** *$D_{\text{align}}$ satisfies the triangle inequality on $\mathcal{G}_{\text{align}}$.*

*Proof.* For each edge slot $a \in \mathcal{A}$, define the coordinate metric

$$d_a(G,H) := H\left(\rho_a^G, \rho_a^H\right), \tag{117}$$

where

$$H(P,Q) := \sqrt{H^2(P,Q)} \tag{118}$$

is the Hellinger distance induced by Eq. (23). By Eq. (28),

$$D_{\text{align}}(G,H) = \frac{1}{\sqrt{m}} \left\| \left(d_a(G,H)\right)_{a\in\mathcal{A}} \right\|_2. \tag{119}$$

Since $H(\cdot,\cdot)$ is a metric on probability measures, Eq. (117) satisfies the triangle inequality for each coordinate:

$$d_a(G,F) \leq d_a(G,H) + d_a(H,F), \qquad a \in \mathcal{A}. \tag{120}$$

Applying Minkowski's inequality to Eq. (119) yields

$$\begin{aligned} D_{\text{align}}(G,F) &= \frac{1}{\sqrt{m}} \left\| \left(d_a(G,F)\right)_{a\in\mathcal{A}} \right\|_2 \\ &\leq \frac{1}{\sqrt{m}} \left\| \left(d_a(G,H) + d_a(H,F)\right)_{a\in\mathcal{A}} \right\|_2 \\ &\leq \frac{1}{\sqrt{m}} \left\| \left(d_a(G,H)\right)_{a\in\mathcal{A}} \right\|_2 + \frac{1}{\sqrt{m}} \left\| \left(d_a(H,F)\right)_{a\in\mathcal{A}} \right\|_2 \\ &= D_{\text{align}}(G,H) + D_{\text{align}}(H,F). \end{aligned} \tag{121}$$

□

**Proposition F.6.** *$D_{\mathrm{geom}}$ is a pseudometric on $\mathcal{G}_{\mathrm{align}}$.*

*Proof.* Define the off-diagonal vectorization map

$$\mathrm{off} : \mathbb{R}^{m\times m} \to \mathbb{R}^{m(m-1)} \tag{122}$$

by stacking all off-diagonal entries in any fixed order. Using the triplet distance-geometry matrix from Eq. (30), define the geometry descriptor

$$\Phi_\Gamma(G) := \frac{1}{\sqrt{m(m-1)}}\,\mathrm{off}(\Gamma_G). \tag{123}$$

Then

$$D_{\mathrm{geom}}(G,H) = \|\Phi_\Gamma(G) - \Phi_\Gamma(H)\|_2\,. \tag{124}$$

Equation (124) is just Euclidean distance between descriptors, so it satisfies the triangle inequality:

$$D_{\mathrm{geom}}(G,F) \le D_{\mathrm{geom}}(G,H) + D_{\mathrm{geom}}(H,F). \tag{125}$$

It is generally only a pseudometric because the descriptor $\Phi_\Gamma(G)$ need not be injective on $\mathcal{G}_{\mathrm{align}}$. □

Equation (33) implies that the same pseudometric can be computed equivalently through the Gram descriptor

$$\Phi_K(G) := \frac{1}{\sqrt{m(m-1)}}\,\mathrm{off}(K_G), \tag{126}$$

since

$$\|\Phi_\Gamma(G) - \Phi_\Gamma(H)\|_2 = \|\Phi_K(G) - \Phi_K(H)\|_2\,. \tag{127}$$

This is the precise sense in which the geometry term may be implemented in Gram form while still representing contextual triplet distance geometry.

**Proposition F.7.** *$D_{\mathrm{metric}}$ satisfies the triangle inequality on $\mathcal{G}_{\mathrm{align}}$.*

*Proof.* By Eqs. (98), (99), and (35),

$$D_{\mathrm{metric}}(G,H) = \left\|\big(D_{\mathrm{align}}(G,H), D_{\mathrm{geom}}(G,H)\big)\right\|_2\,. \tag{128}$$

Using Propositions F.5 and F.6,

$$D_{\mathrm{align}}(G,F) \le D_{\mathrm{align}}(G,H) + D_{\mathrm{align}}(H,F), \tag{129}$$
$$D_{\mathrm{geom}}(G,F) \le D_{\mathrm{geom}}(G,H) + D_{\mathrm{geom}}(H,F). \tag{130}$$

Therefore,

$$\begin{aligned} D_{\mathrm{metric}}(G,F) &= \left\|\big(D_{\mathrm{align}}(G,F), D_{\mathrm{geom}}(G,F)\big)\right\|_2 \\ &\le \left\|\big(D_{\mathrm{align}}(G,H) + D_{\mathrm{align}}(H,F), D_{\mathrm{geom}}(G,H) + D_{\mathrm{geom}}(H,F)\big)\right\|_2 \\ &\le \left\|\big(D_{\mathrm{align}}(G,H), D_{\mathrm{geom}}(G,H)\big)\right\|_2 + \left\|\big(D_{\mathrm{align}}(H,F), D_{\mathrm{geom}}(H,F)\big)\right\|_2 \\ &= D_{\mathrm{metric}}(G,H) + D_{\mathrm{metric}}(H,F), \end{aligned} \tag{131}$$

where the second inequality is Minkowski's inequality in $\mathbb{R}^2$. □

## F.5 Identity of Indiscernibles

The remaining metric axiom is identity of indiscernibles. This is the only step that uses the all-pair slot assumption together with injectivity of the contextual lifting.

**Proposition F.8.** *For $G, H \in \mathcal{G}_{\mathrm{align}}$,*

$$D_{\mathrm{metric}}(G,H) = 0 \quad\Longrightarrow\quad G = H. \tag{132}$$

*Proof.* Assume

$$D_{\text{metric}}(G,H)=0. \tag{133}$$

By Eq. (35),

$$D_{\text{align}}(G,H)=0 \qquad \text{and} \qquad D_{\text{geom}}(G,H)=0. \tag{134}$$

From Eq. (119), the vector of coordinate triplet distances is the zero vector, so for every edge slot $a\in\mathcal{A}$,

$$H\left(\rho_a^G,\rho_a^H\right)=0. \tag{135}$$

Since Hellinger distance is a metric on probability measures, Eq. (135) implies

$$\rho_a^G=\rho_a^H, \qquad a\in\mathcal{A}. \tag{136}$$

We now take marginals of the triplet product measures. By Eq. (82), for every $x\in\Omega_V$,

$$\sum_{y\in\Omega_R}\sum_{z\in\Omega_V}\rho_a^G(x,y,z)=\alpha_{s(a)}^G(x), \tag{137}$$

$$\sum_{x\in\Omega_V}\sum_{z\in\Omega_V}\rho_a^G(x,y,z)=\beta_a^G(y), \qquad y\in\Omega_R, \tag{138}$$

$$\sum_{x\in\Omega_V}\sum_{y\in\Omega_R}\rho_a^G(x,y,z)=\alpha_{t(a)}^G(z), \qquad z\in\Omega_V. \tag{139}$$

Applying the same formulas to $H$ and using Eq. (136), we obtain

$$\alpha_{s(a)}^G=\alpha_{s(a)}^H, \qquad \beta_a^G=\beta_a^H, \qquad \alpha_{t(a)}^G=\alpha_{t(a)}^H, \qquad a\in\mathcal{A}. \tag{140}$$

Since $\mathcal{A}$ contains all directed pairs and $n\geq 2$, every node slot $i\in V$ appears as either the source or the target of at least one edge slot. Hence Eq. (140) implies

$$\alpha_i^G=\alpha_i^H, \qquad i\in V, \tag{141}$$

and, directly from the middle relation in Eq. (140),

$$\beta_a^G=\beta_a^H, \qquad a\in\mathcal{A}. \tag{142}$$

Finally, Corollary F.3 converts equality of lifted measures into equality of the original node and predicate distributions:

$$p_i^G=p_i^H, \qquad i\in V, \tag{143}$$

$$q_a^G=q_a^H, \qquad a\in\mathcal{A}. \tag{144}$$

By Eq. (95), this means $G=H$. □

The converse direction is immediate: if $G=H$ under Eq. (95), then $\alpha_i^G=\alpha_i^H$, $\beta_a^G=\beta_a^H$, and $\rho_a^G=\rho_a^H$ for all slots, so Eqs. (28), (34), and (35) give

$$G=H \implies D_{\text{metric}}(G,H)=0. \tag{145}$$

### F.6 Proof of the Main Metric Theorem

We restate and prove Theorem 3.1.

**Theorem F.9** (Restatement of Theorem 3.1)**.** *Assume the fixed node and edge slots in Eqs.* (1) *and* (2)*, the all-pair setting with $n\geq 2$, the contextual semantic measures in Eqs.* (15) *and* (16)*, the class-specific identity-anchor assumptions from Appendix D.3, and the canonical no-relation measure in Eq.* (17)*. Then the distance $D_{\text{metric}}$ in Eq.* (35) *is a metric on $\mathcal{G}_{\text{align}}$.*

*Proof.* Non-negativity and symmetry follow from Proposition F.4. The triangle inequality follows from Proposition F.7. Identity of indiscernibles follows from Proposition F.8 together with Eq. (145). Therefore, $D_{\text{metric}}$ satisfies all metric axioms on $\mathcal{G}_{\text{align}}$. □

Equation (36) is the squared form of this canonical distance and is therefore a squared metric loss. This is the precise sense in which the proposed graph-level supervision term is metric-compatible on aligned probabilistic 3D scene graphs.

## G Practical Variants

This appendix details the practical variants referenced in Section 3. In particular, it specifies the candidate-edge version of the graph-level objective in Eq. (37), the conversion from sigmoid multi-label relation scores in Eq. (6) to the predicate distribution required by Eq. (20), and the sampling-based estimators associated with the full geometry term in Eq. (34). Throughout, the sigmoid-head conversion and the sampled or weighted geometry estimators are treated as implementational realizations or approximations of the same underlying graph-level supervision principle. By contrast, the candidate-edge variant in Eq. (37) is a separate practical objective defined on a restricted edge-slot domain, while the canonical all-pair distance remains Eq. (35).

### G.1 Candidate Edge Sets

The canonical formulation uses the full directed edge-slot set $\mathcal{A}$ from Eq. (2). In many practical 3D scene graph predictors, however, relation reasoning is restricted to the fixed candidate edge-slot set $E^* \subseteq \mathcal{A}$ introduced in Section 3.3, with cardinality $m^* = |E^*|$.

In the candidate-edge setting, the aligned triplet term and the scene-internal geometry term are obtained from Eqs. (28) and (34) by replacing $\mathcal{A}$ with $E^*$ and replacing $m$ with $m^*$. The corresponding graph-level training loss is the node-augmented objective in Eq. (37). Thus, the candidate-edge setting does not alter the structure of the proposed loss; it only restricts the active edge-slot domain over which the triplet field is instantiated.

For notational convenience, when discussing practical estimators below, we write

$$\mathcal{E} = \begin{cases} \mathcal{A}, & \text{all-pair setting,} \\ E^*, & \text{candidate-edge setting.} \end{cases} \tag{146}$$

Equation (146) lets us describe the all-pair and candidate-edge variants in a unified way without redefining the canonical all-pair distance.

### G.2 Node Block for Candidate Edge Settings

The node block in Eq. (38) is unnecessary in the canonical all-pair setting because every node slot appears in at least one directed edge slot when $n \geq 2$, as used in Appendix F. In contrast, under a candidate-edge set $E^*$, some node slots may be uncovered by the active relation domain.

For a node slot $i \in V$, define its incident candidate-edge set as

$$\mathcal{I}_i(E^*) = \{a \in E^* \mid s(a) = i \text{ or } t(a) = i\}. \tag{147}$$

If

$$\mathcal{I}_i(E^*) = \emptyset, \tag{148}$$

then the lifted node measure $\alpha_i^G$ does not appear in the candidate-edge versions of Eqs. (28) and (34). Consequently, the triplet blocks alone do not constrain the corresponding node distribution at slot $i$.

This is exactly why Eq. (37) includes the additional node block Eq. (38). It restores node-coordinate information independently of the coverage pattern of $E^*$. Therefore, when graph identity is restricted to all node slots and the active relation slots in $E^*$, the square root of Eq. (37) has the same direct-sum metric structure as the all-pair distance proved in Appendix F. It should not be interpreted as a metric on the full all-pair graph if relation distributions outside $E^*$ are still included in graph identity. In other words, the node term is not an auxiliary classification loss; it is the squared form of a Hellinger metric block that keeps the node part of the scene graph representation complete under a restricted edge-slot domain.

**Proposition G.1.** *Assume the same contextual lifting and identity-anchor conditions as in Theorem 3.1. For a fixed candidate edge-slot set $E^* \subseteq \mathcal{A}$, define the restricted aligned graph space*

$$\mathcal{G}_{E^*} = \prod_{i \in V} \Delta^{|\mathcal{C}|} \times \prod_{a \in E^*} \Delta^{|\mathcal{R}_0|}, \tag{149}$$

*with graph identity defined by equality of all node distributions and all predicate distributions on $E^*$. With the degenerate conventions described in Appendix G.6, the square root of Eq.* (37)*,*

$$D_{E^*}(G,H) = \sqrt{D^2_{\text{node}}(G,H) + D^2_{\text{align},E^*}(G,H) + D^2_{\text{geom},E^*}(G,H)}, \tag{150}$$

*is a metric on $\mathcal{G}_{E^*}$.*

*Proof.* The square root of $D^2_{\text{node}}$ is an $\ell_2$ product metric of Hellinger distances between the lifted node measures. By the object-lifting injectivity in Proposition F.1, this node block separates all node distributions. The square root of $D^2_{\text{align},E^*}$ is an $\ell_2$ product metric over the triplet product measures indexed by $E^*$. Therefore, if this block is zero, the predicate lifted measure $\beta_a^G$ equals $\beta_a^H$ for every $a \in E^*$ by taking the predicate marginal of the corresponding triplet product measure. Predicate-lifting injectivity in Proposition F.2 then gives $q_a^G = q_a^H$ for all $a \in E^*$.

The square root of $D^2_{\text{geom},E^*}$ is a Euclidean pseudometric on the off-diagonal triplet distance-geometry descriptor, with the zero convention when $|E^*| < 2$. The direct sum of the node metric, the aligned triplet metric, and this geometry pseudometric satisfies non-negativity, symmetry, and the triangle inequality. If $D_{E^*}(G,H) = 0$, the node and aligned blocks imply equality of all coordinates in Eq. (149). Thus $G = H$ under the restricted fixed correspondence. The converse direction is immediate from the definitions. □

### G.3 Sigmoid Multi-Label Relation Head Conversion

Equation (20) requires a predicate distribution $q_a \in \Delta^{|\mathcal{R}_0|}$. When the model uses a softmax-style predicate head, this distribution is obtained directly. When the model instead uses sigmoid multi-label relation scores as in Eq. (6), we first convert them into a distribution over $\mathcal{R}_0$.

Let $Y_a \subseteq \mathcal{R}$ denote the random set of active positive predicates for edge slot $a$ under the independent Bernoulli model induced by the sigmoid scores:

$$\Pr(r \in Y_a) = s_a(r), \qquad r \in \mathcal{R}. \tag{151}$$

Under this model, the probability of no positive predicate is

$$\Pr(Y_a = \varnothing) = \prod_{r \in \mathcal{R}} \big(1 - s_a(r)\big), \tag{152}$$

and the probability that exactly one positive predicate $r$ is active is

$$\Pr(Y_a = \{r\}) = s_a(r) \prod_{r' \in \mathcal{R} \setminus \{r\}} \big(1 - s_a(r')\big). \tag{153}$$

Because Eq. (20) requires a distribution over the single-label predicate space $\mathcal{R}_0$, we condition the independent Bernoulli model on the event that either no positive predicate is active or exactly one positive predicate is active. Define

$$\mathcal{B}_1 = \{\varnothing\} \cup \big\{\{r\} : r \in \mathcal{R}\big\}. \tag{154}$$

The sigmoid multi-label output is converted into the conditional single-label distribution

$$q_a(\varnothing) = \Pr(Y_a = \varnothing \mid Y_a \in \mathcal{B}_1), \tag{155}$$

$$q_a(r) = \Pr(Y_a = \{r\} \mid Y_a \in \mathcal{B}_1), \qquad r \in \mathcal{R}. \tag{156}$$

Equivalently, after dividing the empty-event and singleton-event probabilities by the common factor $\prod_{r' \in \mathcal{R}} (1 - s_a(r'))$, we obtain the following unnormalized weights:

$$\widetilde{q}_a(\varnothing) = 1, \tag{157}$$

and, for each positive predicate $r \in \mathcal{R}$,

$$\widetilde{q}_a(r) = \frac{s_a(r)}{1 - s_a(r)}. \tag{158}$$

The normalized predicate distribution is then

$$q_a(r) = \frac{\widetilde{q}_a(r)}{\widetilde{q}_a(\varnothing) + \sum_{r' \in \mathcal{R}} \widetilde{q}_a(r')}, \qquad r \in \mathcal{R}_0. \tag{159}$$

This conversion is the exact conditional distribution over $\mathcal{R}_0$ induced by the independent Bernoulli predicate model after conditioning on the event $Y_a \in \mathcal{B}_1$. Multi-positive outcomes are not represented in $\mathcal{R}_0$ because the contextual predicate-measure space and the downstream generator interface use a single predicate label per directed edge slot. Thus, the conversion should be interpreted as a principled single-label projection of a multi-label sigmoid head, rather than as the full independent Bernoulli distribution over all predicate subsets.

This conversion preserves compatibility with common multi-label predicate heads. If several positive predicates receive high sigmoid scores, their masses are distributed according to their Bernoulli odds $s_a(r)/(1 - s_a(r))$. If all positive predicate scores are small, the no-relation mass $q_a(\varnothing)$ remains large because the positive odds are small relative to the no-relation weight in Eq. (157).

### G.4 Sampled Geometry Estimator

The full scene-internal geometry term in Eq. (34) requires all off-diagonal triplet pairs in the active edge-slot set. To unify the all-pair and candidate-edge cases, define the active ordered pair set

$$\mathcal{P}(\mathcal{E}) = \{(a, b) \in \mathcal{E}^2 \mid a \neq b\}, \tag{160}$$

where $\mathcal{E}$ is given by Eq. (146). Using the shorthand

$$\Delta_{ab}(G, H) := \big(K_G[a, b] - K_H[a, b]\big)^2, \qquad (a, b) \in \mathcal{P}(\mathcal{E}), \tag{161}$$

when $\mathcal{P}(\mathcal{E}) \neq \emptyset$, the full active-set geometry term is

$$D^2_{\mathrm{geom},\mathcal{E}}(G, H) = \frac{1}{|\mathcal{P}(\mathcal{E})|} \sum_{(a,b) \in \mathcal{P}(\mathcal{E})} \Delta_{ab}(G, H). \tag{162}$$

In the all-pair case, Eq. (162) coincides with Eq. (34). In the candidate-edge case, it coincides with the geometry block implicitly referenced in Eq. (37).

When $\mathcal{P}(\mathcal{E}) \neq \emptyset$, fix a sample size $T$ with $1 \leq T \leq |\mathcal{P}(\mathcal{E})|$ and draw

$$S \subseteq \mathcal{P}(\mathcal{E}), \qquad |S| = T, \tag{163}$$

uniformly without replacement. The uniform sampled estimator in Eq. (39) is the sample average of the pair discrepancies over $S$. Under this sampling scheme,

$$\mathbb{E}_S \left[\widehat{D}^2_{\mathrm{geom}}(G, H)\right] = D^2_{\mathrm{geom},\mathcal{E}}(G, H). \tag{164}$$

Thus, Eq. (39) is an unbiased estimator of the full active-set geometry term.

The motivation for Eq. (39) is purely computational. It reduces the quadratic cost of evaluating the full pair set while leaving the underlying canonical objective unchanged. This is why the main text explicitly states that the estimator does not change the definition of the full all-pair distance in Eq. (35) or the candidate-edge squared objective in Eq. (37).

### G.5 Weighted Geometry Estimator

Uniform sampling is often adequate, but in highly imbalanced scene graphs the off-diagonal pair set $\mathcal{P}(\mathcal{E})$ may be dominated by pairs whose triplets involve no-relation slots. In such cases, stratified or importance sampling can reduce estimator variance without changing the canonical objective.

Let

$$\pi_{ab} > 0, \qquad (a, b) \in \mathcal{P}(\mathcal{E}), \qquad \sum_{(a,b) \in \mathcal{P}(\mathcal{E})} \pi_{ab} = 1 \tag{165}$$

be a sampling distribution over active ordered edge-slot pairs. If $(a_1, b_1), \ldots, (a_T, b_T)$ are sampled i.i.d. from $\pi$, then the importance-weighted estimator

$$\widehat{D}^2_{\text{geom},\pi}(G, H) = \frac{1}{T} \sum_{t=1}^{T} \frac{\Delta_{a_t b_t}(G, H)}{|\mathcal{P}(\mathcal{E})|\, \pi_{a_t b_t}} \tag{166}$$

satisfies

$$\mathbb{E}\left[\widehat{D}^2_{\text{geom},\pi}(G, H)\right] = D^2_{\text{geom},\mathcal{E}}(G, H). \tag{167}$$

Hence Eq. (166) is an unbiased estimator of the same canonical geometry term.

A useful special case is stratified sampling. For example, one may partition $\mathcal{P}(\mathcal{E})$ into strata according to whether the ordered pair $(a, b)$ is positive-positive, positive-no-relation, or no-relation-no-relation under the corresponding ground-truth relation sets. Sampling rates may then be increased for strata that are rare but informative, while the importance correction in Eq. (166) preserves the target quantity in Eq. (162).

This distinction is important. A weighted or stratified estimator can still be consistent with the canonical metric objective if it is importance-corrected as in Eq. (166). By contrast, a weighted average without such correction should be regarded as a biased optimization surrogate rather than as the squared form of the canonical metric.

### G.6 Numerical Stability

The constructions above are mathematically straightforward but benefit from a few implementation details for numerical robustness.

First, the sigmoid-to-single-label conversion in Appendix G.3 involves Bernoulli odds $s_a(r)/(1 - s_a(r))$. Although sigmoid outputs are mathematically in $(0, 1)$, finite-precision implementations can produce values extremely close to 0 or 1. In practice, we use score clipping only as a numerical approximation to the weights in Eqs. (157) and (158). We set $\varepsilon = 10^{-6}$ and define

$$\bar{s}_a(r) = \min\{1 - \varepsilon, \max\{\varepsilon, s_a(r)\}\}, \qquad r \in \mathcal{R}. \tag{168}$$

The log-space weights are then

$$\log \widetilde{q}_a(\varnothing) = 0, \tag{169}$$

and

$$\log \widetilde{q}_a(r) = \log \bar{s}_a(r) - \log\bigl(1 - \bar{s}_a(r)\bigr), \qquad r \in \mathcal{R}. \tag{170}$$

Normalization in Eq. (159) is implemented through a log-sum-exp denominator. This implementation converges to the exact conditional Bernoulli conversion as $\varepsilon \to 0$, while avoiding undefined logarithms and overflow from near-unit sigmoid scores in finite-precision arithmetic.

Second, the square-root embedding in Eq. (24) is well-defined at zero coordinates, so exact zeros do not create a mathematical problem. In finite precision arithmetic, however, clamping to a small nonnegative floor before the square root can prevent negative values caused by rounding.

Third, when the active edge-slot set is too small to define off-diagonal pairs, the geometry term is set to zero by convention. In particular, if

$$|\mathcal{E}| < 2, \tag{171}$$

then Eq. (162) is undefined as an off-diagonal average, and the implementation uses the degenerate assignment

$$D^2_{\text{geom},\mathcal{E}}(G, H) = 0. \tag{172}$$

If $\mathcal{E} = \emptyset$, the aligned edge average is also undefined. In this degenerate candidate-edge case, we set

$$D^2_{\text{align},\mathcal{E}}(G, H) = 0. \tag{173}$$

Together with Eq. (172), the objective uses only the node block in Eq. (38). This case cannot occur in the canonical all-pair setting under $n \geq 2$. This convention is consistent with the candidate-edge treatment in Section 3.3: if there is no nontrivial edge-domain triplet comparison to compute, the objective reduces to the remaining metric blocks.

Finally, for large scenes, the sampled estimators in Eqs. (39) and (166) should be accumulated in at least standard floating-point precision to avoid cancellation error across many pair discrepancies.

## H Efficient Implementation and Complexity

This appendix presents an efficient implementation of the graph-level objective defined in Section 3, together with its computational and memory cost. The main observation is that the triplet feature in Eq. (26) never needs to be materialized explicitly. All quantities required by the aligned term in Eq. (28) and the geometry term in Eq. (34) can be computed through the factorized triplet affinity in Eq. (27).

### H.1 Dictionary Matrix Construction

The contextual semantic measures from Eqs. (15) and (16) are stored as dictionary matrices. For objects, define

$$M_V \in \mathbb{R}^{|\mathcal{C}|\times|\Omega_V|}, \qquad M_V[c,:] = \mu_V(c), \tag{174}$$

where the row indexed by class $c$ is the contextual object measure from Eq. (15). For predicates, define

$$M_R \in \mathbb{R}^{|\mathcal{R}_0|\times|\Omega_R|}, \qquad M_R[r,:] = \mu_R(r), \tag{175}$$

where the row indexed by $r \in \mathcal{R}_0$ is the contextual relation measure from Eq. (16). In particular, the no-relation row is the canonical measure from Eq. (17).

The matrices in Eqs. (174) and (175) are constructed once from the count tables described in Appendix D and remain fixed during training. They provide a compact matrix form of the contextual semantic dictionary used by the lifting equations in Eqs. (18) to (21).

### H.2 Matrix Form of Semantic Lifting

For a scene with $n$ node slots and an active edge-slot set $\mathcal{E}$, where $\mathcal{E} = \mathcal{A}$ in the all-pair setting and $\mathcal{E} = E^*$ in the candidate-edge setting as in Eq. (146), fix an ordered enumeration $(e_1, \ldots, e_{m_\mathcal{E}})$ of the active edge slots, where $m_\mathcal{E} = |\mathcal{E}|$. Each ordered edge slot in this enumeration is written as $e_\ell = (s(e_\ell), t(e_\ell))$. To match the notation of Section 3.3, we write $D^2_{\text{align},\mathcal{E}}$ and $D^2_{\text{geom},\mathcal{E}}$ for the aligned and geometry blocks evaluated on the active edge-slot set $\mathcal{E}$. When $\mathcal{E} = \mathcal{A}$, these reduce to Eqs. (28) and (34); when $\mathcal{E} = E^*$, they coincide with the corresponding blocks in Eq. (37). For compactness in this appendix, we write $G^{pred} := G_\theta(X)$ for the predicted graph and keep $G^{gt}$ for the canonical ground-truth graph.

We collect the predicted object distributions from Eq. (4) into a matrix

$$P^{pred} \in \mathbb{R}^{n\times|\mathcal{C}|}, \qquad P^{pred}[i,c] = p_i(c), \tag{176}$$

and the predicted predicate distributions from Eq. (5) into

$$Q^{pred} \in \mathbb{R}^{m_\mathcal{E}\times|\mathcal{R}_0|}, \qquad Q^{pred}[\ell,r] = q_{e_\ell}(r), \quad \ell = 1,\ldots,m_\mathcal{E}, \quad r \in \mathcal{R}_0. \tag{177}$$

For the ground-truth graph, define the object target matrix

$$P^{gt} \in \mathbb{R}^{n\times|\mathcal{C}|}, \qquad P^{gt}[i,c] = p_i^{gt}(c), \tag{178}$$

where $p_i^{gt}$ is the canonical object target distribution in Eq. (96). Similarly, define the predicate target matrix

$$Q^{gt} \in \mathbb{R}^{m_\mathcal{E}\times|\mathcal{R}_0|}, \qquad Q^{gt}[\ell,r] = q^{gt}_{e_\ell}(r), \quad \ell = 1,\ldots,m_\mathcal{E}, \quad r \in \mathcal{R}_0, \tag{179}$$

where $q_a^{gt}$ is the canonical predicate target distribution in Eq. (97). Consequently, $Q^{gt}M_R$ implements the ground-truth relation lift in Eq. (21).

Using the dictionary matrices in Eqs. (174) and (175), the contextual semantic lifting in Eqs. (18), (19), (20), and (21) can be written in matrix form as

$$A^{pred} = P^{pred}M_V, \qquad A^{gt} = P^{gt}M_V, \tag{180}$$

and

$$B^{pred} = Q^{pred}M_R, \qquad B^{gt} = Q^{gt}M_R. \tag{181}$$

The $i$-th row of $A^{pred}$ is exactly the lifted measure $\alpha_i^{pred}$ from Eq. (18), and the $i$-th row of $A^{gt}$ is exactly the lifted measure $\alpha_i^{gt}$ from Eq. (19). Likewise, the $\ell$-th row of $B^{pred}$ is the lifted predicate measure $\beta_{e_\ell}^{pred}$ from Eq. (20), and the $\ell$-th row of $B^{gt}$ is the lifted ground-truth measure $\beta_{e_\ell}^{gt}$ from Eq. (21).

### H.3 Hellinger Feature Matrices

Equation (24) defines the Hellinger feature of a lifted measure by an elementwise square root. Applying this row-wise to Eqs. (180) and (181) gives the feature matrices

$$V^{pred} = \sqrt{A^{pred}}, \qquad V^{gt} = \sqrt{A^{gt}}, \tag{182}$$

and

$$U^{pred} = \sqrt{B^{pred}}, \qquad U^{gt} = \sqrt{B^{gt}}, \tag{183}$$

where the square root is applied elementwise.

The row vectors of Eqs. (182) and (183) are exactly the features in Eq. (24). Thus,

$$V^{pred}[i,:] = v_i^{pred}, \qquad V^{gt}[i,:] = v_i^{gt}, \tag{184}$$

and

$$U^{pred}[\ell,:] = u_{e_\ell}^{pred}, \qquad U^{gt}[\ell,:] = u_{e_\ell}^{gt}, \qquad \ell = 1,\dots,m_{\mathcal{E}}. \tag{185}$$

### H.4 Factorized Gram Computation

The key to efficient computation is that the triplet affinity in Eq. (27) factorizes into object and predicate affinities. We therefore first form the object affinity matrices

$$S_V^{pp} = V^{pred}(V^{pred})^\top, \qquad S_V^{gg} = V^{gt}(V^{gt})^\top, \qquad S_V^{pg} = V^{pred}(V^{gt})^\top, \tag{186}$$

and the predicate affinity matrices

$$S_R^{pp} = U^{pred}(U^{pred})^\top, \qquad S_R^{gg} = U^{gt}(U^{gt})^\top, \qquad S_R^{pg} = U^{pred}(U^{gt})^\top. \tag{187}$$

Here the superscripts $pp$, $gg$, and $pg$ denote predicted-predicted, ground-truth-ground-truth, and predicted-ground-truth affinities, respectively.

For implementation, define the source and target index vectors

$$\mathbf{s}, \mathbf{t} \in \{1,\dots,n\}^{m_{\mathcal{E}}}, \qquad \mathbf{s}[\ell] = s(e_\ell), \qquad \mathbf{t}[\ell] = t(e_\ell), \quad \ell = 1,\dots,m_{\mathcal{E}}. \tag{188}$$

The ordering of $\mathbf{s}$ and $\mathbf{t}$ is the same as the ordering of $Q^{pred}$, $Q^{gt}$, $U^{pred}$, and $U^{gt}$.

Using Eq. (27), the within-graph Gram matrix for the predicted graph is computed as

$$K_{pred}[\ell,k] = S_V^{pp}[\mathbf{s}[\ell],\mathbf{s}[k]]\, S_R^{pp}[\ell,k]\, S_V^{pp}[\mathbf{t}[\ell],\mathbf{t}[k]], \qquad \ell,k = 1,\dots,m_{\mathcal{E}}, \tag{189}$$

and the within-graph Gram matrix for the ground-truth graph is

$$K_{gt}[\ell,k] = S_V^{gg}[\mathbf{s}[\ell],\mathbf{s}[k]]\, S_R^{gg}[\ell,k]\, S_V^{gg}[\mathbf{t}[\ell],\mathbf{t}[k]], \qquad \ell,k = 1,\dots,m_{\mathcal{E}}. \tag{190}$$

Equations (189) and (190) are the matrix-form versions of Eq. (27) under the ordering $(e_1,\dots,e_{m_{\mathcal{E}}})$. Thus, $K_{pred}[\ell,k] = K_{G^{pred}}[e_\ell,e_k]$ and $K_{gt}[\ell,k] = K_{G^{gt}}[e_\ell,e_k]$, where $K_G$ is the triplet Gram matrix in Eq. (29).

The aligned cross affinity used in Eq. (28) is obtained from the predicted-ground-truth affinity matrices:

$$A_{\text{cross}}[\ell] = S_V^{pg}[\mathbf{s}[\ell],\mathbf{s}[\ell]]\, S_R^{pg}[\ell,\ell]\, S_V^{pg}[\mathbf{t}[\ell],\mathbf{t}[\ell]], \qquad \ell = 1,\dots,m_{\mathcal{E}}, \tag{191}$$

so that Eq. (28) becomes

$$D^2_{\text{align},\mathcal{E}}(G^{pred},G^{gt}) = \frac{1}{m_{\mathcal{E}}}\sum_{\ell=1}^{m_{\mathcal{E}}}\bigl(1 - A_{\text{cross}}[\ell]\bigr). \tag{192}$$

The expression in Eq. (192) is used when $m_{\mathcal{E}} > 0$; for $m_{\mathcal{E}} = 0$, we use the degenerate assignment in Eq. (173).

For candidate-edge objectives, the node block in Eq. (38) is obtained from the diagonal of $S_V^{pg}$:

$$D^2_{\text{node}}(G^{pred},G^{gt}) = \frac{1}{n}\sum_{i=1}^{n}\bigl(1 - S_V^{pg}[i,i]\bigr). \tag{193}$$

Once $S_V^{pg}$ is available, this block adds only $O(n)$ operations.

Likewise, the geometry term in Eq. (34) is computed by comparing the factorized Gram matrices:

$$D^2_{\mathrm{geom},\mathcal{E}}(G^{pred}, G^{gt}) = \frac{1}{m_{\mathcal{E}}(m_{\mathcal{E}}-1)} \sum_{\ell \neq k} \big(K_{pred}[\ell,k] - K_{gt}[\ell,k]\big)^2, \tag{194}$$

with the convention from Appendix G.6 that the term is set to zero when $m_{\mathcal{E}} < 2$.

Equations (192) and (194) show that both the canonical all-pair loss and its active-set variants can be implemented without ever constructing the tensor-product triplet features $z_a$ from Eq. (26). All required quantities are assembled from node-level and predicate-level affinity matrices.

### H.5 Time and Memory Complexity

We now summarize the computational complexity of the efficient implementation. Let

$$N_{\mathrm{node}} = \sum_{S \in \mathcal{D}_{\mathrm{train}}} |V_S| \tag{195}$$

denote the total number of node occurrences in the training database, and let

$$N_{\mathrm{rel}} = \sum_{S \in \mathcal{D}_{\mathrm{train}}} \sum_{\substack{(i,j) \in V_S^2 \\ i \neq j}} |Y^S_{(i,j)}| \tag{196}$$

denote the total number of positive relation occurrences used by the count rules in Appendix D.

**Offline dictionary construction.** Building the raw count tables from Eqs. (46) to (55) requires one pass over the node occurrences and positive predicate occurrences, so the total counting cost is

$$O(N_{\mathrm{node}} + N_{\mathrm{rel}})\,. \tag{197}$$

The subsequent row normalization and dictionary assembly for the object and predicate tables costs

$$O(|\mathcal{C}|\,|\Omega_V| + |\mathcal{R}_0|\,|\Omega_R|)\,. \tag{198}$$

Thus, the full offline dictionary construction cost is

$$O(N_{\mathrm{node}} + N_{\mathrm{rel}} + |\mathcal{C}|\,|\Omega_V| + |\mathcal{R}_0|\,|\Omega_R|)\,. \tag{199}$$

**Online lifting and Hellinger features.** For a scene with $n$ node slots and $m_{\mathcal{E}}$ active edge slots, the matrix lifts in Eqs. (180) and (181) cost

$$O(n\,|\mathcal{C}|\,|\Omega_V| + m_{\mathcal{E}}\,|\mathcal{R}_0|\,|\Omega_R|)\,, \tag{200}$$

assuming dense matrix multiplication. The elementwise square roots in Eqs. (182) and (183) add

$$O(n\,|\Omega_V| + m_{\mathcal{E}}\,|\Omega_R|) \tag{201}$$

operations.

**Affinity matrices.** Computing the object affinity matrices in Eq. (186) costs

$$O\big(n^2\,|\Omega_V|\big)\,, \tag{202}$$

up to constant factors for the three matrix products. Similarly, computing the predicate affinity matrices in Eq. (187) costs

$$O\big(m_{\mathcal{E}}^2\,|\Omega_R|\big)\,. \tag{203}$$

**Aligned and geometry terms.** Once the affinity matrices are available, the aligned cross affinities in Eq. (191) cost

$$O(m_{\mathcal{E}}), \tag{204}$$

and the full geometry comparison in Eq. (194) costs

$$O(m_{\mathcal{E}}^2). \tag{205}$$

Hence the total full-batch per-scene computation cost is

$$O\Big(n\,|\mathcal{C}|\,|\Omega_V| + m_{\mathcal{E}}\,|\mathcal{R}_0|\,|\Omega_R| + n^2\,|\Omega_V| + m_{\mathcal{E}}^2\,|\Omega_R| + m_{\mathcal{E}}^2\Big). \tag{206}$$

**Sampled geometry.** If the sampled estimator in Eq. (39) is used with a sampled pair set $S$ and the full affinity matrices in Eqs. (186) and (187) are already available, then the geometry accumulation cost becomes

$$O(|S|). \tag{207}$$

This reduces only the final accumulation over edge-slot pairs, not the quadratic cost of constructing the full predicate affinity matrices.

To obtain a subquadratic sampled-geometry implementation, the required predicate affinities can instead be computed on demand, and the required node affinities can be obtained either from the full node affinity matrices or from cached on-demand dot products. The aligned term still requires the diagonal predicted-ground-truth predicate affinities, which cost

$$O(m_{\mathcal{E}}\,|\Omega_R|)\,. \tag{208}$$

The corresponding diagonal node cross-affinities add

$$O(n\,|\Omega_V|) \tag{209}$$

if they are not already available from $S_V^{pg}$. For a sampled pair set $S$, the within-graph predicate affinities required by the sampled geometry estimator cost

$$O(|S|\,|\Omega_R| + |S|)\,, \tag{210}$$

after the node affinity matrices in Eq. (186) have been formed or the corresponding node affinities have been cached. Thus, sampling reduces the quadratic predicate-affinity bottleneck only when the full predicate affinity matrices are not materialized.

**Memory cost.** The dominant dense memory terms are the lifted measure matrices $A^{pred}, A^{gt}, B^{pred}, B^{gt}$, the feature matrices $V^{pred}, V^{gt}, U^{pred}, U^{gt}$, and the affinity or Gram matrices. The asymptotic memory footprint is

$$O\Big(n\,|\Omega_V| + m_{\mathcal{E}}\,|\Omega_R| + n^2 + m_{\mathcal{E}}^2\Big). \tag{211}$$

In the all-pair setting, $m_{\mathcal{E}} = m = n(n-1)$, so the predicate-affinity and geometry stages scale quadratically in the number of directed object-pair slots. In the candidate-edge setting, $m_{\mathcal{E}} = |E^*|$, which reduces both time and memory according to the chosen active edge-slot budget.

Overall, the proposed factorization converts the seemingly high-dimensional triplet objective into a sequence of matrix multiplications and elementwise operations over node-level and predicate-level contextual features. This is the precise computational content of the efficiency claim summarized in Section 3.4.